\documentclass[final]{cvpr}

\usepackage{times}
\usepackage{epsfig}
\usepackage{graphicx}
\usepackage[utf8]{inputenc} 
\usepackage[T1]{fontenc}    
\usepackage{url}            
\usepackage{booktabs}       
\usepackage{amsfonts}       
\usepackage{nicefrac}       
\usepackage{microtype}      
\usepackage{xcolor}

\usepackage{amsmath,bm}
\usepackage{amssymb}
\usepackage{enumitem}
\usepackage{subfigure}
\usepackage{multirow}
\usepackage{caption}

\usepackage{float}

\newcommand{\x}{{\bf x}}
\newcommand{\y}{{\bf y}}

\newcommand{\z}{{\bf z}}
\newcommand{\w}{{\bf w}}

\newcommand{\bphi}{\bm{\phi}}
\newcommand{\btheta}{\bm{\theta}}
\newcommand{\EE}{{\mathbb{E}}}

\newcommand\blfootnote[1]{%
  \begingroup
  \renewcommand\thefootnote{}\footnote{#1}%
  \addtocounter{footnote}{-1}%
  \endgroup
}

\newtheorem{theor}{Remark}

\usepackage[pagebackref=true,breaklinks=true,colorlinks,bookmarks=false]{hyperref}

\def\cvprPaperID{6303} 
\def\confYear{CVPR 2021}

\setlist[itemize]{leftmargin=4.mm}

\begin{document}

\title{Dual Contradistinctive Generative Autoencoder}

\author{Gaurav Parmar$^{1}$\thanks{empty}\qquad Dacheng Li$^{1}$\footnotemark[1]\qquad Kwonjoon Lee$^{2}$\footnotemark[1]\qquad Zhuowen Tu$^{2}$  \\
$^1$Carnegie Mellon University\qquad 
$^2$UC San Diego\qquad \\
\texttt{\{gparmar,dacheng2\}@andrew.cmu.edu}\qquad \texttt{\{kwl042,ztu\}@ucsd.edu} \\
}


\twocolumn[{%
\renewcommand\twocolumn[1][]{#1}%
\vspace{-1mm}
\maketitle
\vspace{-1mm}
\begin{center}
    \vspace{-0.3in}
    \includegraphics[width=0.8\linewidth]{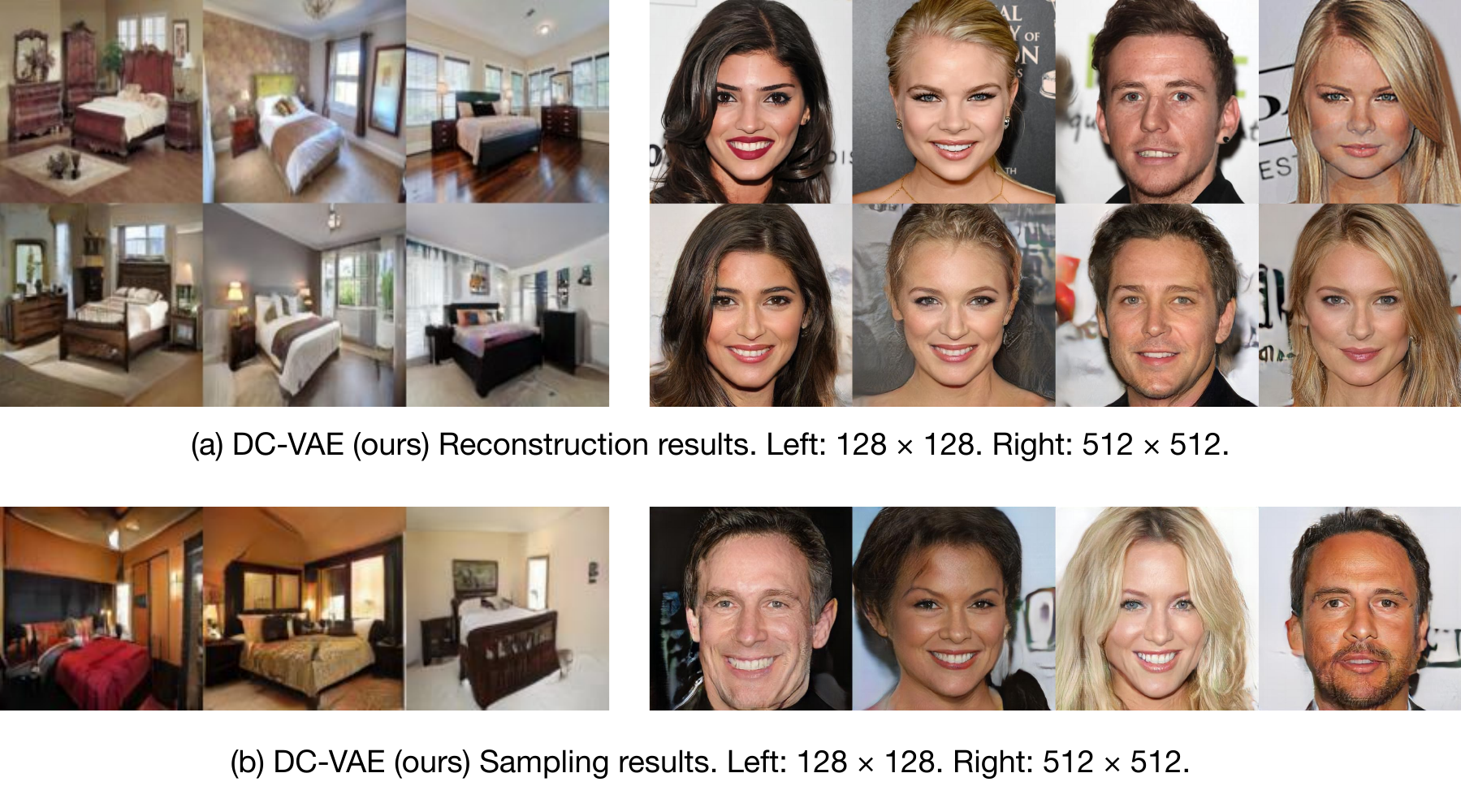}
    \vspace{-5mm}
    \captionof{figure}{DC-VAE Reconstruction (top) and Sampling (bottom) on LSUN Bedroom \cite{yu15lsun} at resolution $128 \times 128$ (left) and CelebA-HQ \cite{karras2018progressive} at resolution $512 \times 512$ (right).}
    \label{teaser}
    
\end{center}%
}]


\begin{abstract}
We present a new generative autoencoder model with dual contradistinctive losses 
to improve generative autoencoder that performs simultaneous inference (reconstruction) and synthesis (sampling). 
Our model, named dual contradistinctive generative autoencoder (DC-VAE), integrates an instance-level discriminative loss (maintaining the instance-level fidelity for the reconstruction/synthesis) with a set-level adversarial loss (encouraging the set-level fidelity for the reconstruction/synthesis), both being contradistinctive. 
Extensive experimental results by DC-VAE across different resolutions including $32 \times 32$, $64 \times 64$, $128 \times 128$, and $512 \times 512$ are reported. The two contradistinctive losses in VAE work harmoniously in DC-VAE leading to a significant qualitative and quantitative performance enhancement over the baseline VAEs without architectural changes.
State-of-the-art or competitive results among generative autoencoders for image reconstruction, image synthesis, image interpolation, and representation learning are observed.
DC-VAE is a general-purpose VAE model, applicable to a wide variety of downstream tasks in computer vision and machine learning.
\end{abstract}

\vspace{-12mm}
\section{Introduction}
\label{sec:intro}

Tremendous progress has been made in deep learning for  the development of various learning frameworks \cite{krizhevsky2012imagenet,he2016deep,goodfellow2014generative,vaswani2017attention}. 
Autoencoder (AE) \cite{lecun1987modeles,hinton1994autoencoders} aims to compactly represent and faithfully reproduce the original input signal by concatenating an encoder and a decoder in an end-to-end learning framework.
The goal of AE is to make the encoded representation semantically {\em efficient} and {\em sufficient} to reproduce the input signal by its decoder.
Autoencoder's generative companion, variational autoencoder (VAE) \cite{kingma2013auto}, additionally learns a variational model for the latent variables to capture the underlying sample distribution. 

The key objective for a generative autoencoder is to maintain two types of fidelities:
(1) an {\em instance-level fidelity} to make the reconstruction/synthesis faithful to the individual input data sample, and (2) a {\em set-level fidelity} to make the reconstruction/synthesis of the decoder faithful to the entire input data set. The VAE/GAN algorithm \cite{VAEGAN} combines a reconstruction loss with an adversarial loss. However, the result of VAE/GAN is sub-optimal, as shown in Table \ref{table:cifar_ablation}.\blfootnote{* indicates equal contribution}


The pixel-wise reconstruction loss in the standard VAE \cite{kingma2013auto} typically results in blurry images with degenerated semantics. A possible solution to resolving the above conflict lies in two aspects: (1) turning the measure in the pixel space into induced feature space that is more semantically meaningful; (2) changing the L2 distance (per-pixel) into a \textit{learned} \textit{instance-level} distance function for the entire image (akin to generative adversarial networks which learn \textit{set-level} distance functions). Taking these two steps allows us to design an instance-level classification loss that is aligned with the adversarial loss in the GAN model enforcing set-level fidelity. Motivated by the above observations, we develop a new generative autoencoder model with dual contradistinctive losses by adopting a discriminative loss performing instance-level classification (enforcing the instance-level fidelity), which is rooted in metric learning \cite{kulis2012metric} and contrastive learning \cite{hadsell2006dimensionality,wu2018unsupervised,infoNCE}. Combined with the adversarial losses for the set-level fidelity, both terms are formulated in the induced feature space performing contradistinction:
(1) the instance-level contrastive loss considers each input instance (image) itself as a class, and (2) the set-level adversarial loss treats the entire input set as a positive class. We name our method dual contradistinctive generative autoencoder (DC-VAE) and make the following contributions.
\begin{itemize}
 \setlength\itemsep{0mm}
 \setlength{\itemindent}{0mm}
    \item We develop a new algorithm, dual contradistinctive generative autoencoder (DC-VAE), by combining instance-level and set-level classification losses in the  VAE framework,
     and systematically show the significance of these two loss terms in DC-VAE.   
    \item The effectiveness of DC-VAE is illustrated in a number of tasks, including image reconstruction, image synthesis, image interpolation, and representation learning by reconstructing and sampling images across different resolutions including $32 \times 32$, $64 \times$, $128 \times 128$, and $512 \times 512$.
    \item Under the new loss term, DC-VAE attains a significant performance boost over the competing methods without architectural change, making it a general-purse model applicable to a variety of computer vision tasks. DC-VAE helps greatly reducing the performance gap for image synthesis between the baseline VAE to the competitive GAN models.
\end{itemize}

\section{Related Work}
\label{sec:related}
Related work can be roughly divided into three categories: (1) generative autoencoder, (2) deep generative model, and (3) contrastive learning.

\noindent\textbf{Generative autoencoder}. Variational autoencoder (VAE) \cite{kingma2013auto} points to an exciting direction of generative models by developing an Evidence Lower BOund (ELBO) objective \cite{higgins2017beta, ding2020guided}. However, the VAE reconstruction/synthesis is known to be blurry. To improve the image quality, a sequence of VAE based models have been developed \cite{VAEGAN,dumoulin2017adversarially,huang2018introvae,brock2018large,zhang2019perceptual}. VAE/GAN \cite{VAEGAN} adopts an adversarial loss to improve the quality of the image, but its output for both reconstruction and synthesis (new samples) is still unsatisfactory.
IntroVAE \cite{huang2018introvae} adds a loop from the output back to the input and is able to attain image quality that is on par with some modern GANs in some aspects. However, its full illustration for both reconstruction and synthesis is unclear. PGA \cite{zhang2019perceptual} adds a constraint to the latent variables.


\noindent\textbf{Deep generative model}. Pioneering works of \cite{tu2007learning,NCE} 
alleviate the difficulty of learning densities by approximating likelihoods via classification (real (positive) samples vs. fake  (pseudo-negative or adversarial) samples). Generative adversarial network (GAN) \cite{goodfellow2014generative} builds on neural networks and amortized sampling (a decoder network that maps a noise into an image). The subsequent development after GAN \cite{DCGAN,WGAN,gulrajani2017improved,karras2018progressive,gong2019autogan,dumoulin2017adversarially,donahue2017bigan} has led to a great leap forward in building decoder-based generative models.
It has been widely observed that the adversarial loss in GANs contributes significantly to the improved quality of image synthesis.  Energy-based generative models \cite{pmlr-v5-salakhutdinov09a,xie2016theory,jin2017introspective,lee2018wasserstein}
--- which aim to directly model data density --- are making a steady improvement for a simultaneously generative and discriminative single model.

\noindent\textbf{Contrastive learning}. From another angle, contrastive learning \cite{hadsell2006dimensionality,wu2018unsupervised,he2020momentum,chen2020simple} has lately shown its particular advantage in unsupervised training of CNN features. It overcomes the limitation in unsupervised learning where class label is missing by turning each image instance into one class. Thus, the softmax function in the standard discriminative classification training can be applied. Contrastive learning can be connected to metric learning \cite{bromley93, chopra2005, chechik2010}. 

In this paper, we aim to improve VAE \cite{kingma2013auto} by introducing a contrastive loss \cite{infoNCE} to 
address instance-level fidelity between the input and the reconstruction in the induced feature space.
Unlike in self-supervised representation learning methods \cite{infoNCE,he2020momentum,chen2020simple}, where self-supervision requires generating a transformed input (via data augmentation operations), the reconstruction naturally fits into the contrastive term that encourages the matching between the reconstruction and the input image instance, while pushing the reconstruction away from the rest of the images in the entire training set. Thus, the instance-level and set-level contradistinctive terms collaborate with each to encourage the high fidelity of the reconstruction and synthesis. In Figure \ref{fig:CF10_abl}, we systematically show the significance of with and without the instance-level and the set-level contradistinctive terms. In addition, we explore multi-scale contrastive learning via two schemes in Section \ref{sec:multi_scale}: 1) deep supervision for contrastive learning in different convolution layers, and 2) patch-based contrastive learning for fine-grained data fidelity. In the experiments, we show competitive results for the proposed DC-VAE in a number of benchmarks for three tasks, including image synthesis, image reconstruction, and representation learning.

\section{Preliminaries: VAE and VAE/GAN}

\paragraph{Variational autoencoder (VAE)} Assume a given training set $S=\{\x_i\}_{i=1}^n$ where each $\x_i \in \mathbb{R}^m$. We suppose that each $\x_i$ is sampled from a  generative process $p(\x|\z)$. In the literature, vector $\z$ refers to latent variables. In practice, latent variables $\z$ and the generative process $p(\x|\z)$ are unknown. 
The objectives of a variational autoencoder (VAE) \cite{kingma2013auto} is to simultaneously train an inference network $q_{\bphi}(\z|\x)$ and a generator network $p_{\btheta}(\x|\z)$. In VAE \cite{kingma2013auto}, the inference network is a neural network that outputs parameters for Gaussian distribution $q_{\bphi}(\z|\x)=\mathcal{N}(\mu_{\bphi}(\x),\Sigma_{\bphi}(\x))$.
The generator is a deterministic neural network $f_{\btheta}(\z)$ 
parameterized by $\btheta$. Generative density $p_{\btheta}(\x|\z)$ is assumed to be subject to a Gaussian distribution: 
$p_{\btheta}(\x|\z)=\mathcal{N}(f_{\btheta}(\z), \sigma^2 I)$. These models can be trained by minimizing the \textbf{negative} of evidence lower bound (ELBO) in Eq. (\ref{eq:ELBO}) below.
\begin{equation}
\begin{aligned}
    &\mathcal{L}_\text{ELBO}(\btheta, \bphi; \x) = \\
    &-\EE_{\z \sim q_{\bphi}(\z|\x)} [\log(p_{\btheta}(\x|\z))] + KL[q_{\bphi}(\z|\x) || p(\z)]
\end{aligned}
\label{eq:ELBO}
\end{equation}
where $p(\z)$ is the prior, which is assumed to be $\mathcal{N}(0, I)$. The first term $-\EE_{q_{\bphi}(\z|\x)} [\log(p_{\btheta}(\x|\z))]$ reduces to standard pixel-wise reconstruction loss $\EE_{q_{\bphi}(\z|\x)}[||\x-f_{\btheta}(\z)||_2^2]$ (up to a constant) due to the Gaussian assumption. The second term is the regularization term, which prevents the conditional $q_{\bphi}(\z|\x)$ from deviating from the Gaussian prior $\mathcal{N}(0, I)$.
The inference network and generator network are jointly optimized over training samples by:
\begin{equation}
    \min_{\btheta, \bphi} \mathop{\mathbb{E}}_{\x \sim p_{\text{data}}(\x)} \mathcal{L}_\text{ELBO}(\btheta, \bphi; \x).
\label{eq:VAE}
\end{equation}
where $p_{\text{data}}$ is the distribution induced by the training set $S$.

VAE has an elegant formulation. However, it relies on a pixel-wise reconstruction loss, which is known not ideal to be reflective of perceptual realism \cite{Johnson2016Perceptual, pix2pix2017}, often resulting in blurry images. From another viewpoint, it can be thought of as using a kernel density estimator (with an isotropic Gaussian kernel) in the pixel space. Although allowing efficient training and inference, such a non-parametric approach is overly simplistic for modeling the semantics and perception of natural images.

\paragraph{VAE/GAN} Generative adversarial networks (GANs) \cite{goodfellow2014generative} and its variants \cite{DCGAN}, on the other hand, are shown to be producing highly realistic images. The success was largely attributed to learning a fidelity function (often referred to as a discriminator) that measures how realistic the generated images are. This can be achieved by learning to contrast (classify) the set of training images with the set of generated images \cite{tu2007learning, NCE, goodfellow2014generative}.

VAE/GAN \cite{VAEGAN} augments the ELBO objective (Eq. (\ref{eq:VAE})) with the GAN objective. Specifically, the objective of VAE/GAN consists of two terms, namely the modified ELBO (Eq. (\ref{eq:VAEGAN_ELBO})) and the GAN objective. To make the notations later consistent, we now define the set of given training images as $S=\{\x_{i}\}_{i=1}^n$ in which a total number of $n$ unlabeled training images are present. 
For each input image $\x_i$, the modified ELBO computes the reconstruction loss in the \textit{feature space} of the discriminator instead of the pixel space:
\begin{equation}
\small
\begin{aligned}
    &\mathcal{L}_\text{ELBO}(\btheta, \bphi, D; \x_i) = \\
    &\EE_{\z \sim q_{\bphi}(\z|\x_i)} [||F_D(\x_i)-F_D(f_{\btheta}(\z))||_2^2] + KL[q_{\bphi}(\z|\x_i) || p(\z)]
\end{aligned}
\label{eq:VAEGAN_ELBO}
\end{equation}
where $F_D(\cdot)$ denotes the feature embedding from the discriminator $D$. Feature reconstruction loss (also referred to as perceptual loss), similar to that in style transfer \cite{Johnson2016Perceptual}.
The modified GAN objective considers both reconstructed images (latent code from $q_{\bphi}(\z|\x)$) and sampled images (latent code from the prior $p(\z)$) as its fake samples:
\begin{equation}
\small
\begin{aligned}
    \mathcal{L}_\text{GAN}(\btheta, \bphi, D; \x_i) &= \log D(\x_i) + \EE_{\z \sim p(\z)}\log (1 - D(f_{\btheta}(\z))\\
    &   + \EE_{\z \sim q_{\bphi}(\z|\x_i)}\log (1 - D(f_{\btheta}(\z)).
\end{aligned}
\label{eq:VAEGAN_GAN}
\end{equation}
The VAE/GAN objective becomes:
\begin{equation}
    \min_{\btheta, \bphi}\max_D \sum_{i=1}^n \left[ \mathcal{L}_\text{ELBO}(\btheta, \bphi,D; \x_i) + \mathcal{L}_\text{GAN}(\btheta, \bphi, D; \x_i) \right].
\label{eq:VAEGAN_obj}
\end{equation}

\begin{figure*}[!htp]
\begin{center}
\scalebox{0.9}{
\begin{tabular}{cc}
\hspace{-2mm}
\includegraphics[width=0.9\textwidth]{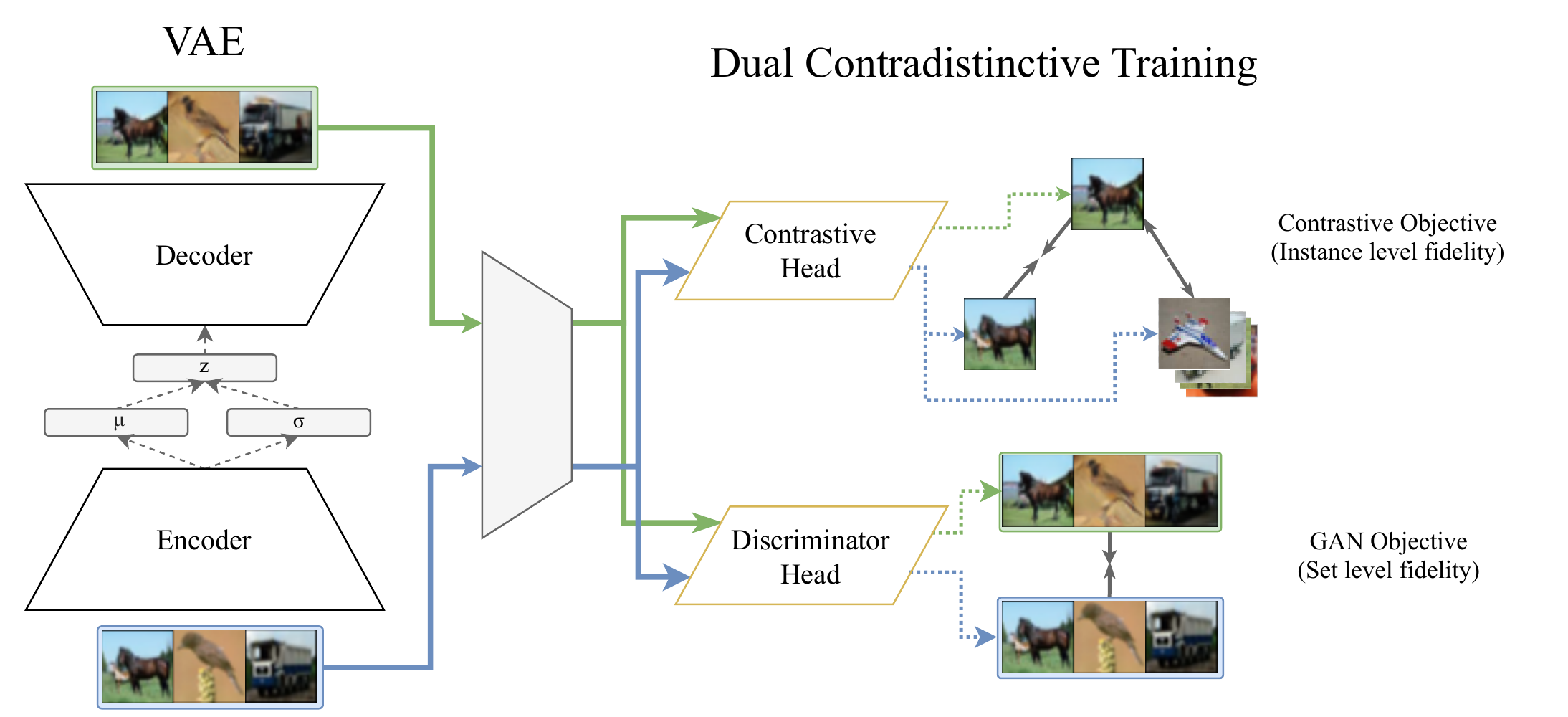} 
\hspace{-2mm}
\end{tabular}
}
\vspace{-2mm}
\caption{\textbf{Model architecture} for the proposed DC-VAE algorithm.}
\label{fig:model}
\vspace{-6mm}
\end{center}
\end{figure*}

\vspace{-2mm}
\section{Dual contradistinctive generative autoencoder (DC-VAE)}
\vspace{-2mm}

Here we want to address a question: {\em Is the degeneration of the synthesized images by VAE always the case once the decoder is joined with an encoder?} Can the problem be remedied by using a more informative loss?

Although improving the image qualities of VAE by integrating a set-level contrastive loss (GAN objective of Eq. (\ref{eq:VAEGAN_GAN})), VAE/GAN still does not accurately model instance-level fidelity. Inspired by the literature on instance-level classification \cite{exemplar-svm}, approximating likelihood by classification \cite{tu2007learning},  and contrastive learning \cite{hadsell2006dimensionality,wu2018unsupervised,he2020momentum},
we propose to model instance-level fidelity by contrastive loss (commonly referred to as InfoNCE loss) \cite{infoNCE}. In DC-VAE, we perform the following minimization and loosely call each term a loss.
\begin{equation}\label{eqn:infonce}
\begin{aligned}
     &\mathcal{L}_\text{instance}(\btheta, \bphi, D; i, \{\x_{j}\}_{j=1}^n) \triangleq \\
     &-\mathbb{E}_{\z \sim q_{\bphi}(\z|\x_i)}\left[\log \frac{e^{h\left(\x_i, f_{\btheta}(\z)\right)}}{\sum_{j=1}^{n} e^{h\left(\x_{j}, f_{\btheta}(\z)\right)}}\right],
\end{aligned}
\end{equation}
where $i$ is an index for a training sample (instance), $\{\x_{j}\}_{j=1}^n$ is the union of positive samples and negative samples, $h(\x,\y)$ is the critic function that measures compatibility between $\x$ and $\y$. Following the popular choice from \cite{he2020momentum}, $h(\x,\y)$ is the cosine similarity between the embeddings of $\x$ and $\y$: $h(\x,\y)=\frac{F_D(\x)^\top F_D(\y)}{||F_D(\x)||_2||F_D(\y)||_2}$. Note that unlike in contrastive self-supervised learning methods \cite{infoNCE,he2020momentum,chen2020simple} where two views (independent augmentations) of an instance constitutes a positive pair, an input instance $\x_i$ and its reconstruction $f_{\btheta}(\z)$ comprises a positive pair in DC-VAE. Likewise, the reconstruction $f_{\btheta}(\z)$ and any instance that is not $\x_i$ can be a negative pair.

To bridge the gap between the instance-level contrastive loss (Eq. (\ref{eqn:infonce})) and log-likelihood in ELBO term (Eq. (\ref{eq:ELBO})), we observe the following connection.
\begin{theor} 
\label{th:nce_theorem}
(From \cite{ma-collins-2018-noise,pmlr-v97-poole19a}) The following objective is minimized, \textit{i.e.}, the optimal critic $h$ is achieved, when $h(f_{\btheta}(\z),\x)=\log p(\x|\z)+c(\x)$ where $c(\x)$ is any function that does not depend on $\z$.

\begin{equation}\label{eqn:INCE}
     I_{\text{NCE}}\triangleq\EE_{\x_1,\cdots \x_K} \mathbb{E}_{i}[\mathcal{L}_\text{instance}(\btheta, \bphi, D; i, \{\x_{j}\}_{j=1}^n)].
\end{equation}

\end{theor}

It can be seen from \cite{ma-collins-2018-noise,pmlr-v97-poole19a} that the contrastive loss of Eq. (\ref{eqn:infonce}) \textit{implicitly} estimates the log-likelihood $\log p_{\theta}(\x|\z)$ required for the evidence lower bound (ELBO). Hence, we modify the ELBO objective of Eq. (\ref{eq:ELBO}) as follows and name it as \textit{implicit} ELBO (IELBO):

\begin{equation}
\begin{aligned}
    &\mathcal{L}_\text{IELBO}(\btheta, \bphi, D; \x_i) = \\ &\mathcal{L}_\text{instance}(\btheta, \bphi, D; i, \{\x_{j}\}_{j=1}^n) + KL[q_{\bphi}(\z|\x_i) || p(\z)].
\end{aligned}
\label{eq:DCAE_ELBO}
\end{equation}

Finally, the combined objective for the proposed DC-VAE algorithm becomes:
\begin{equation}
    \min_{\btheta, \bphi}\max_D \sum_{i=1}^n \left[ \mathcal{L}_\text{IELBO}(\btheta, \bphi, D; \x_i) + \mathcal{L}_\text{GAN}(\btheta, \bphi, D; \x_i) \right].
\label{eq:DCAE_obj}
\end{equation}

\begin{figure*}[!hpt]
\vspace{-2mm}
\begin{center}
  \includegraphics[width=0.85\linewidth]{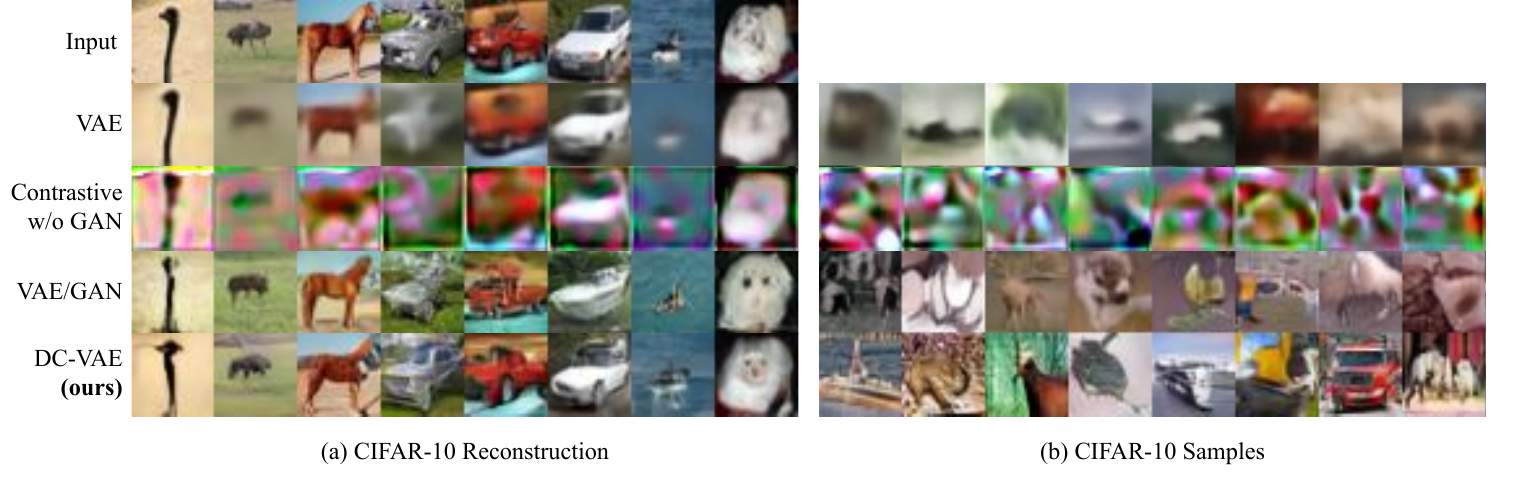}
\vspace{-1mm}
\caption{\small
    Qualitative results of CIFAR-10 \cite{cifar10} images (resolution \(32 \times 32\)) for experiments in Table \ref{table:cifar_ablation} \cite{cifar10}.
}
\vspace{-6mm}
\label{fig:CF10_abl}
\end{center}
\end{figure*}

The definition of $\mathcal{L}_\text{GAN}$ follows Eq. (\ref{eq:VAEGAN_GAN}).
Note here we also consider the term in Eq. (\ref{eq:VAEGAN_GAN}) as contrasdistinctive since it tries to minimize the difference/discriminative classification between the input (``real'') image set and the reconstructed/generated (``fake'') image set. Below we highlight the significance of the two contradistinctive terms. Figure \ref{fig:model} shows the model architecture.

\subsection{Understanding the loss terms}

\noindent \textbf{Instance-level fidelity.} The first item in Eq. (\ref{eq:DCAE_ELBO}) is an instance-level fidelity term encouraging the reconstruction to be as close as possible to the input image while being different from all the rest of the images. A key advantage of the contrastive loss in Eq. (\ref{eq:DCAE_ELBO}) over the standard reconstruction loss in Eq. (\ref{eq:VAEGAN_ELBO}) is its relaxed and background instances aware formulation.
    In general, the reconstruction in Eq. (\ref{eq:VAEGAN_ELBO}) wants a perfect match between the reconstruction and the input, whereas the contrastive loss in Eq. (\ref{eq:DCAE_ELBO}) requests for being the most similar one among the training samples. This way, the contrastive loss becomes more cooperative with less conflict to the GAN loss, compared with the reconstruction loss.
    The introduction of the contrastive loss results in a significant improvement over VAE and VAE/GAN.
    
    We further explain the difference between reconstruction and contrastive loss based on the input $\x$ and it's reconstruction $f_{\theta}(\z)$. To simplify the notation, we use $\x$ instead of the output layer feature $F_D(\x)$ (shown in Eq. \ref{eq:VAEGAN_GAN})) for the illustration purpose.
The reconstruction loss enforces the similarity between the reconstructed image and the input image $\min ||\x-f_{\theta}(\z)||$  while the GAN loss computes an adversarial loss $\min \max_{\w} \log (\frac{1}{1+\exp\{-\w \cdot \x\}}) + \log (1-\frac{1}{1+\exp \{-\w \cdot f_{\theta}(\z)\}})$. $\w$ refers to the classifier parameter. The reconstruction loss term enforces pixel-wise/feature matching between input and the reconstruction, while the GAN loss encourages the reconstruction and input discriminatively non-separable; the two are measured in different ways resulting in a conflict. Our contrastive loss on the other hand, is also a discriminative term, it can be viewed as $\min -\log \frac{\exp\{(\x \cdot f_{\theta}(\z))\}}{\sum_{j=1}^n \exp\{(\x_j \cdot  f_{\theta}(\z))\}}$. To compare the reconstruction loss with the contrastive loss: the former wants to have an exact match between the reconstruction with the input, whereas the later is more relaxed to be ok if no exact match but as the closest one amongst all the training samples.

In other words, the reconstruction wants a perfect match for the instance-level fidelity whereas the contrastive loss is asking for being the most similar one among the given training samples. Using the contrastive loss gives more room and creates less conflict with the GAN loss.

\noindent \textbf{Set-level fidelity.} The second item in Eq. (\ref{eq:DCAE_obj}) is a set-level fidelity term encouraging the entire set of synthesized images to be non distinguishable from the input image set. Having this term (Eq. (\ref{eq:VAEGAN_GAN})) is still important since the instance contrastive loss alone (Eq. (\ref{eq:DCAE_obj})) will also lead to a degenerated situation: the input image and its reconstruction can be projected to the same point in the new feature space, but without a guarantee that the reconstruction itself lies on the valid ``real'' image manifold.

As shown in Figure \ref{fig:CF10_abl} and Table \ref{table:cifar_ablation} for the comparison with and without the individual terms in Eq. (\ref{eq:DCAE_obj}). We observe evident effectiveness of the proposed DC-VAE combining both the instance-level fidelity term (Eq. (\ref{eqn:infonce})) and the set-level fidelity term (Eq. (\ref{eq:VAEGAN_GAN})), compared with VAE (using pixel-wise reconstruction loss without the GAN objective), VAE-GAN (using feature reconstruction loss and the GAN objective), and VAE contrastive (using contrastive loss but without the GAN objective).



In the experiments, we show that both terms required to achieve faithful reconstruction (captured by InfoNCE loss) with perceptual realism (captured by the GAN loss).

\subsection{Multi scale contrastive learning \label{sec:multi_scale}}
Inspired by \cite{lee2015deeply}, we utilize information from feature maps at different scales. In addition to contrasting on the last layer of $D$ in Equation \ref{eq:DCAE_obj}, we add contrastive objective on $f_l(\z)$ where $f_l$ is some function on top of an intermediate layer $l$ of D. We do it in two different ways. 
\begin{enumerate}
 \setlength\itemsep{0mm}
 \setlength{\itemindent}{0mm}
    \item Deep supervision: We use 1$\times$1 convolution to reduce the dimension channel-wise, and use a linear layer to obtain $f_l$.
    \item Local patch: We use a random location across channel at layer $l$ (size: 1$\times$1$\times$d, where d is the channel depth).
\end{enumerate}
The intuition for the second is that in a convolutional neural network, one location at a feature map corresponds to a receptive area (patch) in the original image. Thus, by contrasting locations across channels in the same feature maps, we are encouraging the original image and the reconstruction to image have locally similar content, while encouraging them to have locally dissimilar content in other images. We use deep supervision for initial training, and add local patch after certain iterations. 

\section{Experiments}
\label{sec:experiments}

\subsection{Implementation}
\noindent{\bf Datasets} To validate our method, we train our method on several different datasets --- CIFAR-10 \cite{cifar10}, STL-10 \cite{STL10}, CelebA \cite{CelebA}, CelebA-HQ \cite{karras2018progressive}, and LSUN bedroom \cite{yu15lsun}. See the appendix for more detailed descriptions.

\noindent{\bf Network architecture}  For $32 \times 32$ resolution, we design the encoder and decoder subnetworks of our model in a similar way to the discriminator and generator found through neural architecture search in AutoGAN \cite{gong2019autogan}. For the higher resolution experiments ($128\times 128$ and $512 \times 512$ resolution), we use Progressive GAN \cite{karras2018progressive}
as the backbone. Network architecture diagram is available in the appendix.

\noindent{\bf Training details} The number of negative samples for contrastive learning is 8096 for all datasets (analysis of this hyperparameter is provided in supplementary material). The latent dimension for the VAE decoder is 128 for CIFAR-10, STL-10, and 512 for CelebA, CelebA-HQ and LSUN Bedroom. Learning rate is 0.0002 with Adam parameters of $(\beta_1,\beta_2)=(0.0, 0.9)$ and a batch size of 128 for CIFAR-10 and STL-10. For CelebA, CelebA-HQ, LSUN Bedroom datasets, we use the optimizer parameters given in \cite{karras2018progressive}. The contrastive embedding dimension used is 16 for each of the experiments.

\vspace{-2mm}
\subsection{Ablation Study}
\vspace{-2mm}

\begin{table}[!htp]
\vspace{-2mm}
\centering
\caption{\small \textbf{Ablation studies on CIFAR-10} for the proposed DC-VAE algorithm. We follow \cite{Johnson2016Perceptual} and measure perceptual distance in an relu$4\_3$ layer of a pretrained VGG network. $\downarrow$ means lower is better. $\uparrow$ means higher is better.} 
\label{table:cifar_ablation}
\vspace{-2mm}
\scalebox{0.7}{
\setlength{\tabcolsep}{3.0pt}
\begin{tabular}{@{}lcccccccc@{}}
\hline
\toprule
\textbf{Method} & \begin{tabular}{c} \textbf{FID}$\downarrow$/\textbf{IS}$\uparrow$\\ \textbf{Sampling} \end{tabular} & \begin{tabular}{c} \textbf{FID}$\downarrow$/\textbf{IS}$\uparrow$ \\ \textbf{Reconstruction} \end{tabular} & \begin{tabular}{c} \textbf{Pixel}$\downarrow$ \\ \textbf{Distance} \end{tabular} & \begin{tabular}{c} \textbf{Perceptual}$\downarrow$ \\ \textbf{Distance} \end{tabular} \\
\midrule
VAE & 115.8 / 3.8 & 108.4 / 4.3 & \textbf{21.8} & 65.8 \\
VAE/GAN & 39.8 / 7.4 & 29.0 / 7.6 & 62.7 & 57.2 \\
VAE-Contrastive & 240.4 / 1.8 & 242 / 1.9 & 53.6 & 104.2 \\
DC-VAE  & \textbf{17.9 / 8.2} & \textbf{21.4 / 7.9} & 45.9 & \textbf{52.9} \\
\bottomrule
\hline
\end{tabular}
}
\vspace{-2mm}
\end{table}

To demonstrate the necessity of the GAN loss (Eq. \ref{eq:VAEGAN_GAN}) and contrastive loss (Eq. \ref{eq:DCAE_ELBO}), we conduct four experiments with the same backbone. These experiments are: VAE (No GAN, no Contrastive), VAE/GAN (with GAN, no Contrastive), VAE-Contrastive (No GAN, with Contrastive, and ours (With GAN, with Contrastive). Here, GAN denotes Eq. \ref{eq:VAEGAN_GAN}, and Contrastive denotes Eq. \ref{eq:DCAE_ELBO}.

\begin{table}[!htp]
\vspace{-1mm}
\caption{\small \textbf{Comparison on CIFAR-10 and STL-10.} Average Inception scores (IS) \cite{salimans2016improved} and FID scores \cite{FID}. Results derived from \cite{gong2019autogan}. Table style based \cite{lee2019meta}. \textsuperscript{$\dagger$}Result from \cite{aneja2020ncpvae}.  \textsuperscript{$\ast$}Result from \cite{dieng2019prescribed}.} 
\label{tab:IS_CIFAR}
\vspace{-5mm}
\begin{center}
\begin{small}
\scalebox{0.9}
{%
\begin{tabular}{@{}lc@{}cc@{}c@{}cc@{}}
\hline
\toprule
& \phantom{a} & \multicolumn{2}{c}{\textbf{CIFAR-10}} & \phantom{ab} & \multicolumn{2}{c}{\textbf{STL-10}} \\
\cmidrule{3-4} \cmidrule{6-7}
\textbf{Method} && \textbf{IS}$\uparrow$ & \textbf{FID}$\downarrow$  && \textbf{IS}$\uparrow$ & \textbf{FID}$\downarrow$  \\
\hline
\textit{Methods based on GAN:} && & && & \\
DCGAN \cite{DCGAN} && 6.6  & - && - & - \\
ProbGAN \cite{he2019probgan} && 7.8 & 24.6 && 8.9  & 46.7 \\
WGAN-GP ResNet \cite{gulrajani2017improved} &&  7.9  & - && - & - \\
RaGAN \cite{jolicoeur2018relativistic} && -  &  23.5 && -  & - \\
SN-GAN \cite{miyato2018spectral}  && 8.2  & 21.7  && 9.1  & 40.1  \\
MGAN \cite{hoang2018mgan} && 8.3  & 26.7 && - & - \\
Progressive GAN \cite{karras2018progressive} && \textbf{8.8}  & - && - & - \\
Improving MMD GAN \cite{wang2019improving} && 8.3 & 16.2 && \textbf{9.3} & 37.6 \\
PULSGAN \cite{PUGAN} && -  &  22.3 && -  & - \\
AutoGAN \cite{gong2019autogan} && 8.6  & \textbf{12.4} && 9.2  & \textbf{31.0} \\
\hline
\textit{Methods based on VAE:} && & && & \\
VAE && 3.8  & 115.8  && -  & -  \\
VAE/GAN && 7.4 & 39.8  && -  & -  \\
VEEGAN\textsuperscript{$\ast$} \cite{veegan2017} && - & 95.2  && - & - \\
WAE-GAN \cite{WAE}                         && - & 93.1      && - & - \\
NVAE\textsuperscript{$\dagger$} \cite{vahdat2020NVAE} Sampling && - & 50.8  && -  & -\\
NVAE\textsuperscript{$\dagger$} \cite{vahdat2020NVAE} Reconstruction && -  & \textbf{2.67}  && -  & -\\
DC-VAE Sampling (ours) && \textbf{8.2} & \textbf{17.9}  && 8.1  & \textbf{41.9}  \\
DC-VAE Recon. (ours) && 7.9  & 21.4 && \textbf{8.4}  & 43.6   \\
\bottomrule
\hline
\end{tabular}
\vspace{-2mm}
}
\end{small}

\end{center}
\vspace{-4mm}
\end{table}

\noindent{\bf Qualitative analysis} From Figure \ref{fig:CF10_abl},  we see that without GAN and contrastive, images are blurry; Without GAN, the contrastive head can classify images, but not on the image manifold; Without Contrastive, reconstruction images are on the image manifold because of the discriminator, but they are different from input images. These experiments show that it is necessary to combine both instance-level and set-level fidelity, and in a contradistinctive manner. 

\noindent{\bf Quantitative analysis} In Table \ref{table:cifar_ablation} we observe the same trend. VAE generates blurry images; thus the FID/IS (Inception Score) is not ideal. VAE-Contrastive does not generate images on the natural manifold; thus FID/IS is poor. VAE/GAN combines set-level and instance-level information. However the L2  objective is not ideal; thus the FID/IS is sub-optimal. For both reconstruction and sampling tasks, DC-VAE generates high fidelity images and has a favorable FID and Inception score. This illustrates the advantange of having a contradistinctive objective on both set level and instance level. To measure the faithfulness of the reconstructed image we compute the pixelwise L2 distance and the perceptual distance (\cite{Johnson2016Perceptual}). 
For the pixel distance, VAE has the lowest value because it directly optimizes this distance during training; our pixel-wise distance is better than VAE/GAN and VAE-Contrastive. For perceptual distance, our method outperforms other three, which confirms that using contrastive learning helps reconstruct images semantically.

\vspace{-2mm}
\subsection{Comparison to existing generative models}
\vspace{-2mm}

\begin{figure}[htp!]
\vspace{-1mm}
\begin{center}
    \includegraphics[width=1.0\linewidth]{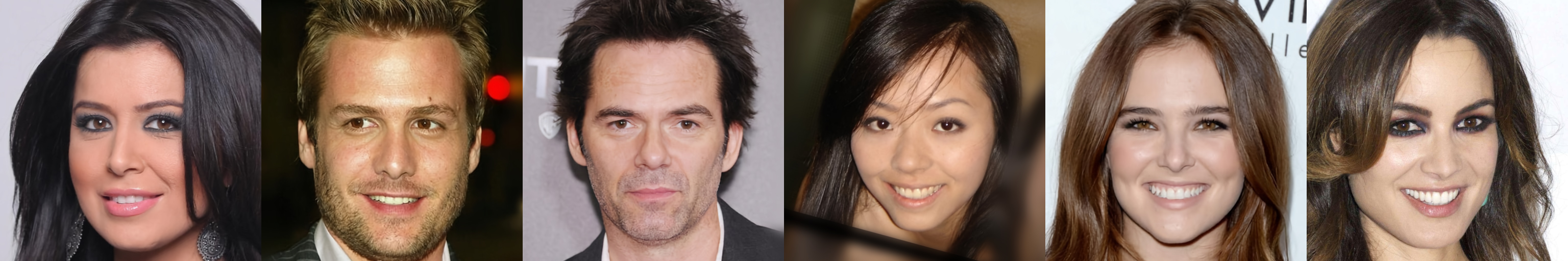}\\
    {\small (a) Input Image ($1024 \times 1024$)} \\
    \includegraphics[width=1.0\linewidth]{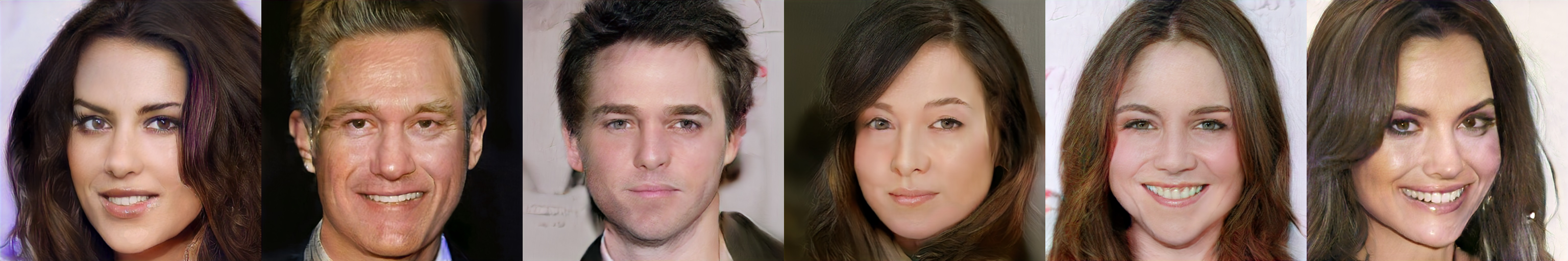}\\
    {\small (b) IntroVAE Reconstruction ($1024 \times 1024$)} \\
    \includegraphics[width=1.0\linewidth]{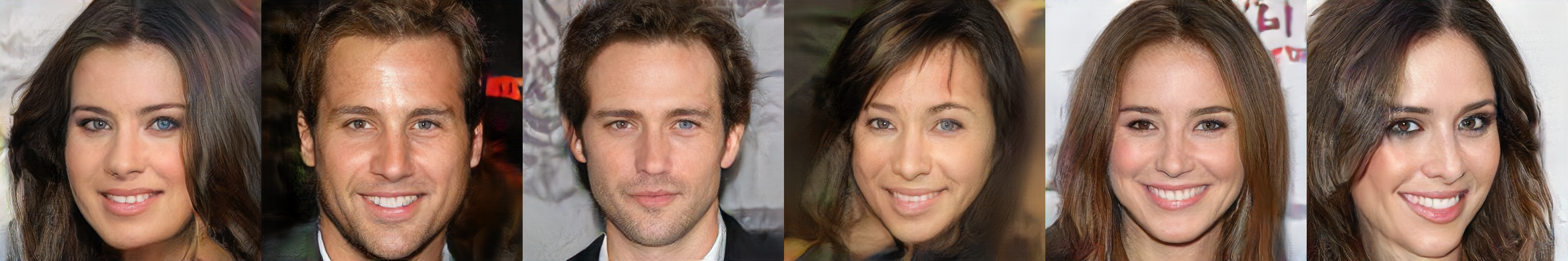}\\
    {\small (c) DC-VAE Reconstruction (ours, $512 \times 512$)}
\end{center}
\vspace{-5mm}
\caption{\small Comparison of DC-VAE (resolution \(512 \times 512\)) with IntroVAE \cite{huang2018introvae} (resolution \(1024 \times 1024\)). Zoom in for a better visualization.}
\label{fig:celebaHQ_intro_compare}
\vspace{-3mm}
\end{figure}

Table \ref{tab:IS_CIFAR} gives a comparison of quantitative measurement for CIFAR-10 and STL-10 dataset. In general, there is a large difference in terms of FID and IS between GAN family and VAE family of models. Our model has state-of-the-art results in VAE family, and is comparable to state-of-the-art GAN models on CIFAR-10. Similarly Tables \ref{tab:singlebest}, \ref{table:celeba}, and \ref{table:celebahq} show that DC-VAE is able to generate images that are comparable to GAN based methods even on higher resolution datasets such as LSUN Bedrooms, CelebA, CelebA-HQ. Our method achieves state-of-the-art results on these datasets among VAE-based methods which focus on building better architectures. Figure \ref{fig:celebaHQ_intro_compare} and Table \ref{tab:recon} show that our model yields more faithful reconstructions compared to existing state-of-the-art generative auto-encoder methods.


\begin{table}[!htp]
\vspace{-1mm}
\centering
\caption{\small \textbf{Quality of Image generation (FID) comparison on LSUN Bedrooms.} \textsuperscript{$\dagger$}128$\times$128 resolution. \textsuperscript{$\ddagger$}256$\times$256 resolution. $\downarrow$ means lower is better.}
\vspace{-5mm}
\label{tab:singlebest}
\begin{center}
\scalebox{0.9}{
\setlength{\tabcolsep}{2.0pt}
\begin{tabular}{@{}lcccccc@{}}
\hline
\toprule
\textbf{Method} & \begin{tabular}{c} \textbf{FID}$\downarrow$ \end{tabular} & \begin{tabular}{c} \textbf{FID}$\downarrow$ \end{tabular} \\
 & (Sampling) & (Reconstruction) \\ 
\midrule
Progressive GAN\textsuperscript{$\ddagger$} \cite{karras2018progressive} 
& \textbf{8.3} & -\\
SNGAN\textsuperscript{$\dagger$} \cite{miyato2018spectral} (from \cite{chen2019self}) & 16.0 &-\\
SSGAN\textsuperscript{$\dagger$}\cite{chen2019self} & 13.3 & -\\
StyleALAE\textsuperscript{$\ddagger$} \cite{pidhorskyi2020adversarial} & 17.13 & 15.92 \\
DC-VAE \textsuperscript{$\dagger$} (ours) & 14.3 & \textbf{10.57} \\
\bottomrule
\hline
\end{tabular}
}
\end{center}
\vspace{-5mm}
\end{table}


\begin{table}
\vspace{-2mm}
\centering
\begin{tabular}{cc}
    \begin{minipage}[t]{1.5in}
                \caption{\small \textbf{FID comparison on CelebA-HQ for 256x256 resolution.} $\downarrow$ means lower is better.}
               \label{table:celebahq} 
        \scalebox{0.8}{
\begin{tabular}{@{}lc@{}}
\hline
\toprule
\textbf{Method} & \textbf{FID}$\downarrow$ \\
\midrule
StyleALAE \cite{pidhorskyi2020adversarial} 
& 19.21 \\
NVAE \cite{vahdat2020NVAE} (from \cite{aneja2020ncpvae}) & 40.26 \\ 
NCP-VAE \cite{aneja2020ncpvae} & 24.69 \\
DC-VAE (ours)
& \textbf{15.81}\\

\bottomrule
\hline
\end{tabular}
}
\end{minipage}

&
    \begin{minipage}[t]{1.5in}
            \caption{\small \textbf{FID comparison on CelebA.}  \textsuperscript{$\ast$}64$\times$64 resolution. \textsuperscript{$\dagger$}128$\times$128 resolution. $\downarrow$ means lower is better.} 
        \label{table:celeba}
        \scalebox{0.62}{
\begin{tabular}{@{}lc@{}}
\hline
\toprule
\textbf{Method} & \textbf{FID}$\downarrow$ \\
\midrule
\textit{Methods based on GAN:}  \\
PresGAN\textsuperscript{$\ast$} \cite{dieng2019prescribed} 
& 29.1  \\
LSGAN \cite{mao2017least} (from \cite{glann2019}) 
& 53.9    \\
COCO-GAN \textsuperscript{$\dagger$} \cite{lin2019coco} & \textbf{5.7} \\

ProGAN\textsuperscript{$\dagger$} \cite{karras2018progressive} (from \cite{lin2019coco}) & 7.30  \\
\hline
\textit{Methods based on VAE:}  \\
VEE-GAN\textsuperscript{$\dagger$} \cite{veegan2017} (from \cite{dieng2019prescribed}) 
& 46.2   \\
WAE-GAN\textsuperscript{$\ast$} \cite{WAE}
& 42    \\
DC-VAE\textsuperscript{$\dagger$} (ours) Reconstruction
& \textbf{14.3}  \\
DC-VAE\textsuperscript{$\dagger$} (ours) Sampling
& \textbf{19.9}  \\

\bottomrule
\hline
\end{tabular}
}
\end{minipage}

\end{tabular}
\vspace{-5mm}
\end{table}

\begin{figure*}[!htp]
\vspace{-1mm}
\begin{center}
  \includegraphics[width=0.475\linewidth]{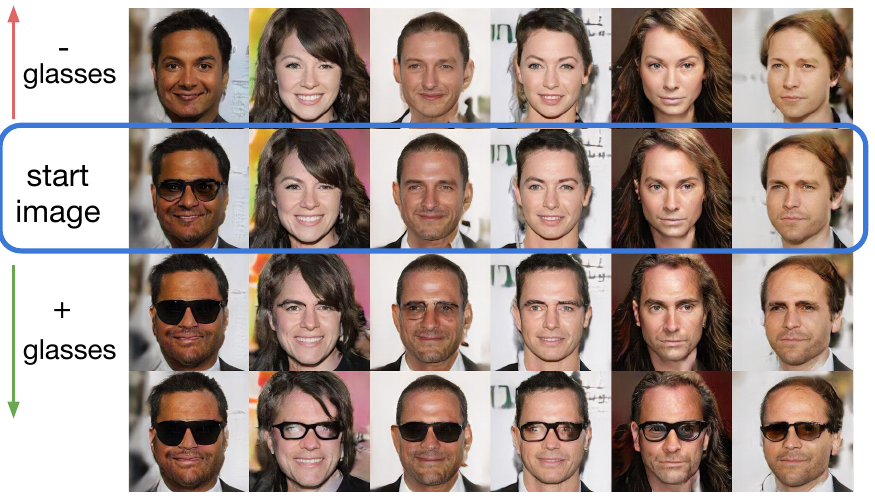}
  \includegraphics[width=0.475\linewidth]{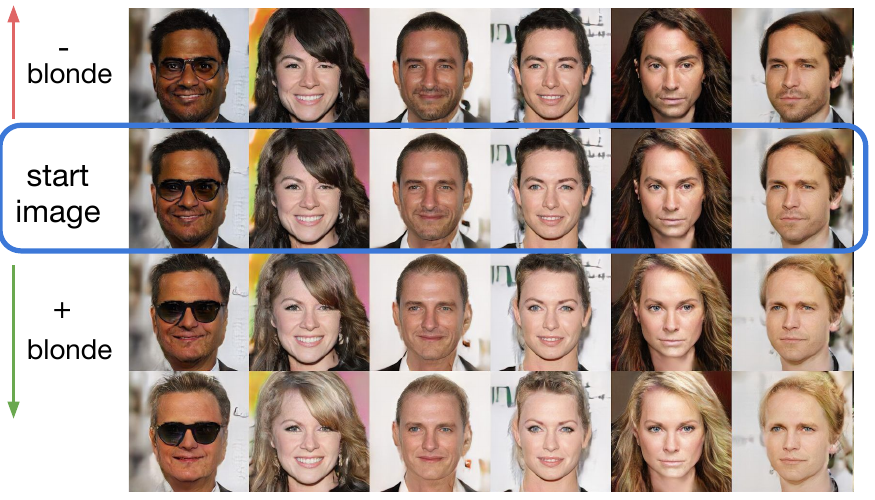}\\
  \includegraphics[width=0.475\linewidth]{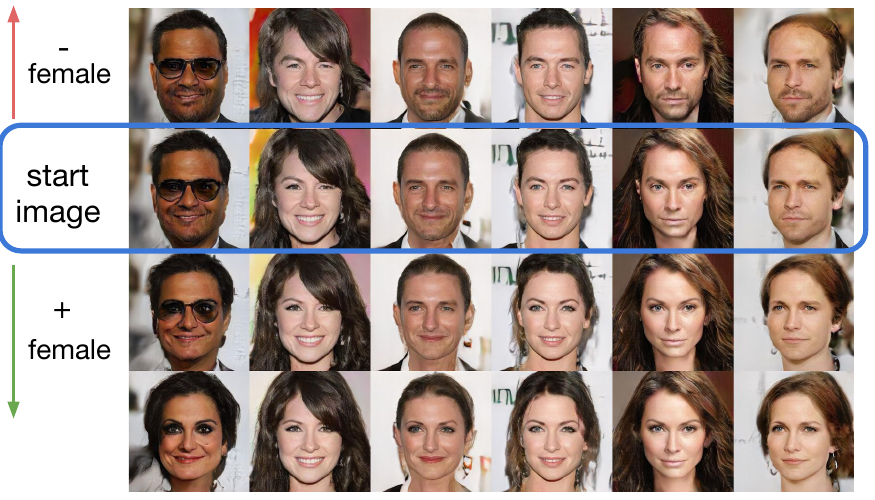}
  \includegraphics[width=0.475\linewidth]{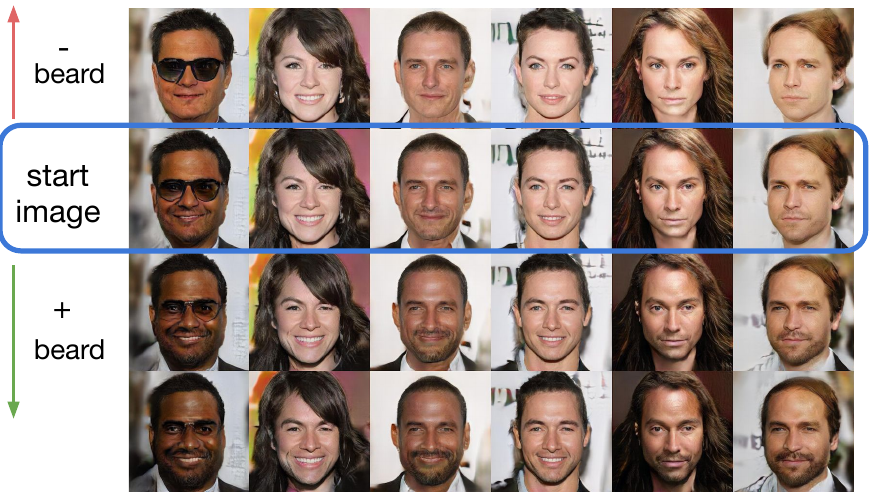}\\
  \includegraphics[width=0.475\linewidth]{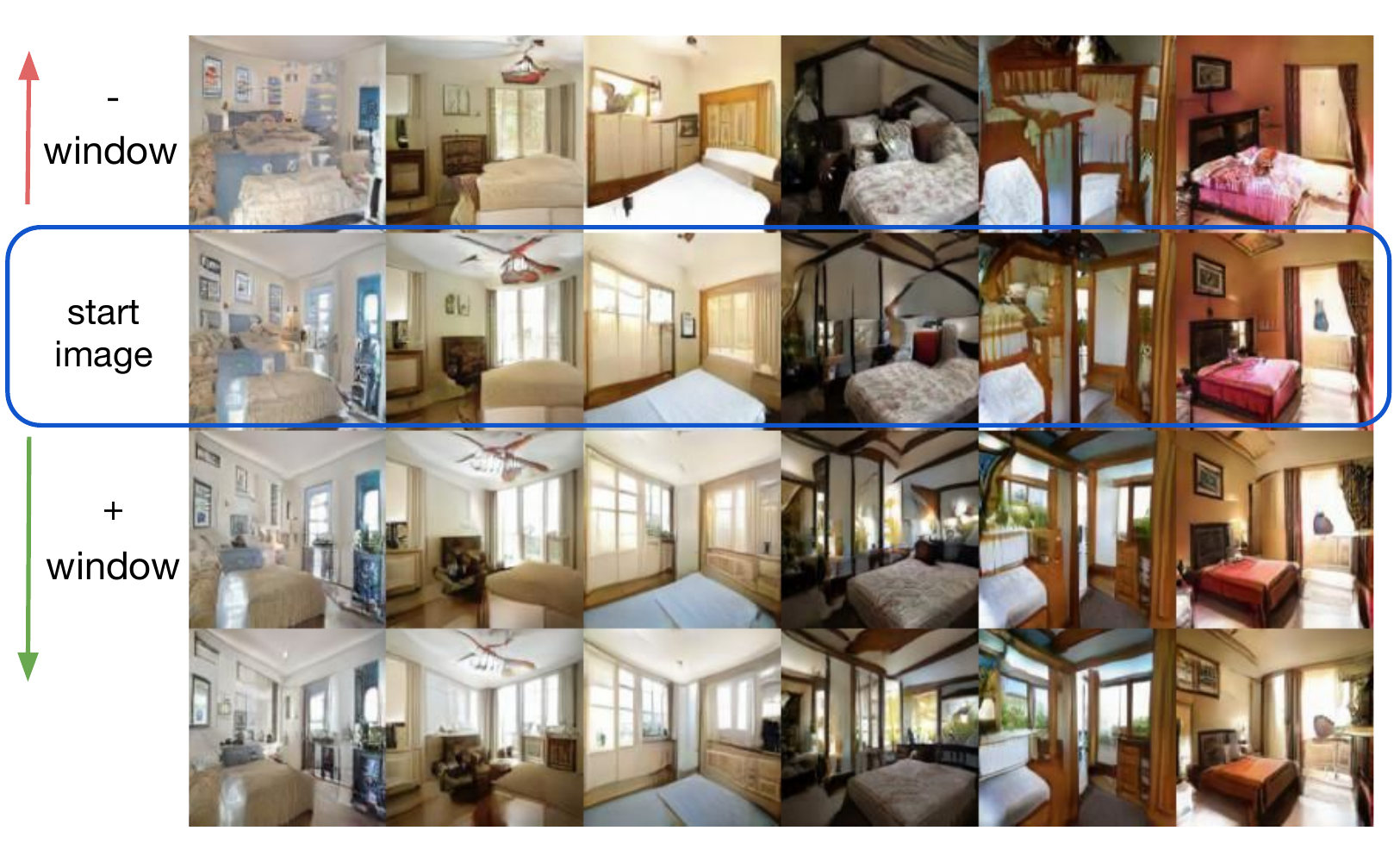}
  \hspace{3mm}
  \includegraphics[width=0.45\linewidth]{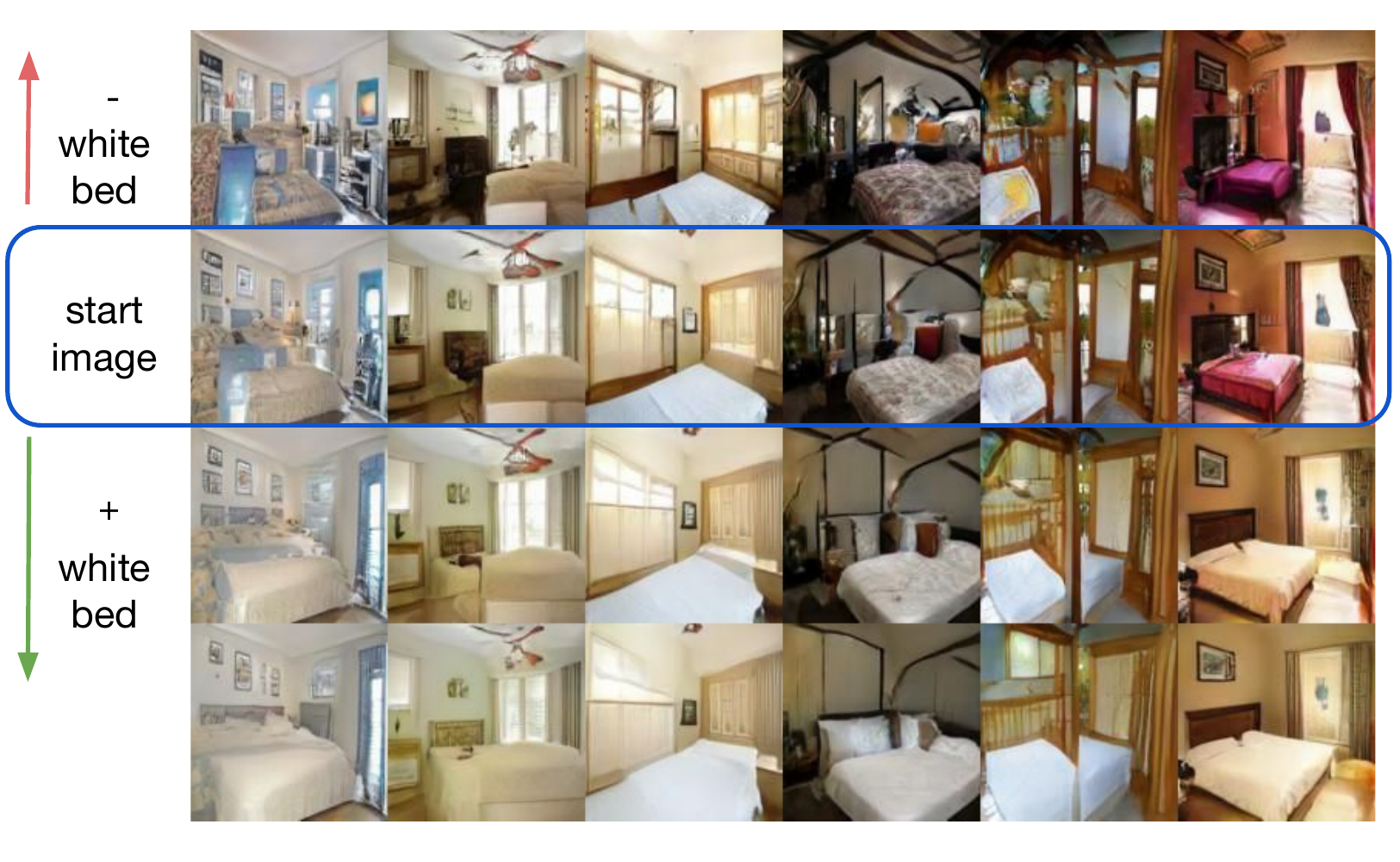}\\
  \includegraphics[width=0.8\linewidth]{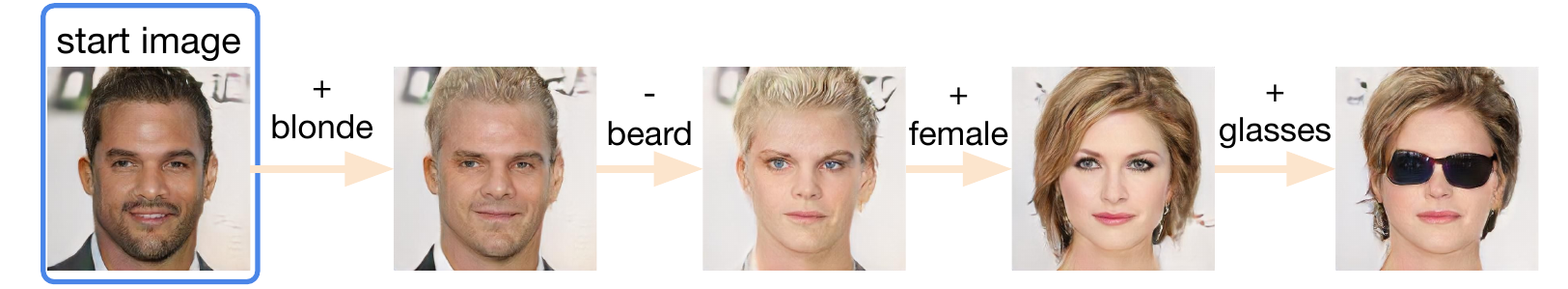}
  
  \caption{\small
    Latent traversal on CelebA-HQ \cite{karras2018progressive} (resolution \(512 \times 512\)) and LSUN Bedroom \cite{yu15lsun} (resolution \(128 \times 128\)) and example image editing on CelebA-HQ \cite{karras2018progressive} image. (Zoom in for a better visualization.). 
}
\vspace{-3mm}
\label{fig:CLB_editing}
\end{center}
\end{figure*}

\begin{figure*}[!htp]
\begin{center}
\begin{tabular}{cc}
  \includegraphics[height=5.8cm]{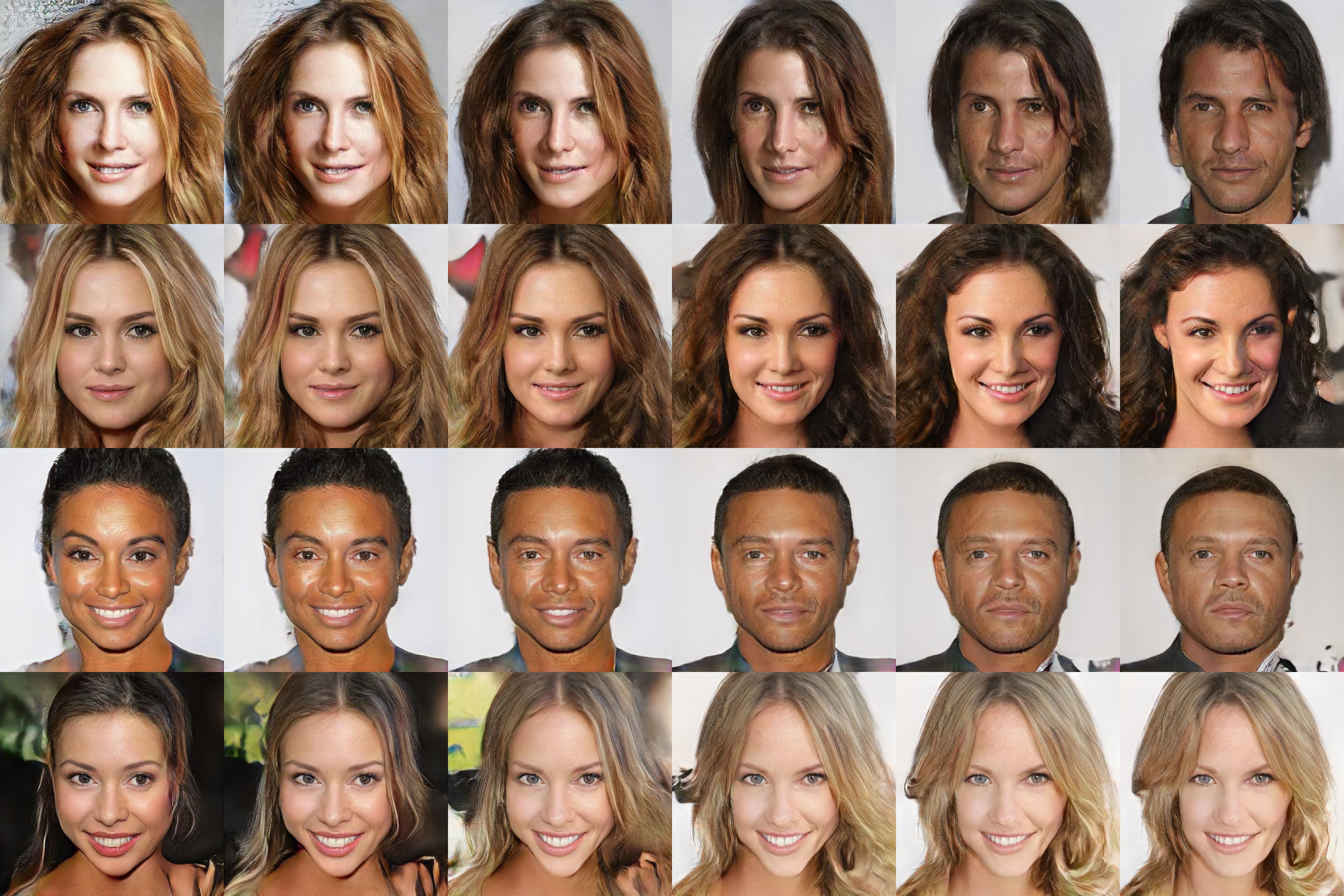} &
  \includegraphics[height=5.8cm]{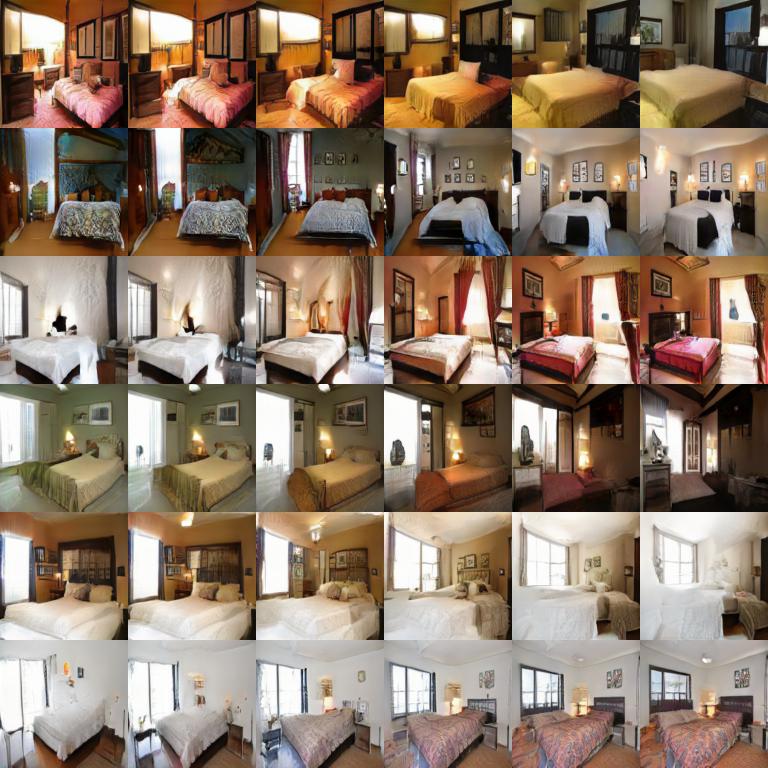}\\
\vspace{-7mm}
\end{tabular}
\caption{\small
    Interpolation results generated by DC-VAE (ours) on CelebA-HQ \cite{karras2018progressive} images (\(512 \times 512\), left) and LSUN Bedroom \cite{yu15lsun} images (\(128 \times 128\), right). (Zoom in for a better visualization.)
}
\vspace{-3mm}
\label{fig:CLB_inter}
\end{center}
\vspace{-1mm}
\end{figure*}

\vspace{-2mm}
\subsection{Latent Space Representation: Image and style interpolation}
\vspace{-2mm}
We further validate the effectiveness of DC-VAE for representation learning. One benefit of having an AE/VAE framework compared with just a decoder as in GAN \cite{goodfellow2014generative} is to be able to directly obtain the latent representation from the input images. The encoder and decoder modules in VAE allows us to readily perform image/style interpolation by mixing the latent variables of different images and reconstruct/synthesize new ones. We demonstrate qualitative results on image interpolation (Fig. \ref{fig:CLB_inter}), style interpolation and image editing (Fig. \ref{fig:CLB_editing}) (method used for this is outlined in the supplementary materials) . We directly use the trained DC-VAE model without disentanglement learning \cite{karras2019style}.
We also quantitatively compare the
 latent space disentanglement through the perceptual path length (PPL) \cite{karras2019style} (Table \ref{tab:PPL}). We observe that DC-VAE learns a more disentangled latent space representation than the backbone Progressive GAN \cite{karras2018progressive} and StyleALAE \cite{pidhorskyi2020adversarial} that use a much more capable StyleGAN \cite{karras2019style} backbone.

\vspace{-2mm}
\subsection{Latent Space Representation: Classification}
\vspace{-2mm}

\begin{table}[h!]
\vspace{1mm}
\begin{center}
\caption{\small \textbf{Comparison to prior VAE-based representation learning methods.} Classification error on MNIST dataset. $\downarrow$: lower is better. 95 \% confidence intervals are from 5 trials. Results derived from \cite{ding2020guided}.}
\label{tab:classification-mnist-methods}
\scalebox{0.7}{
\begin{tabular}{@{}lccc@{}}
\hline
\toprule
\textbf{Method} & {$d_\z = 16$ $\downarrow$} &  {$d_\z = 32$ $\downarrow$}  &  {$d_\z = 64$ $\downarrow$} \\
\hline
VAE \cite{kingma2013auto}
            & 2.92\%$\pm$0.12 & 3.05\%$\pm$0.42 & 2.98\%$\pm$0.14\\
$\beta$-VAE($\beta$=2) \cite{higgins2017beta} 
            & 4.69\%$\pm$0.18 & 5.26\%$\pm$0.22 & 5.40\%$\pm$0.33 \\
FactorVAE($\gamma$=5) \cite{kim2018disentangling} 
            & 6.07\%$\pm$0.05 & 6.18\%$\pm$0.20 & 6.35\%$\pm$0.48 \\
$\beta$-TCVAE ($\alpha$=1,$\beta$=5,$\gamma$=1) \cite{chen2018isolating} 
            & 1.62\%$\pm$0.07 & 1.24\%$\pm$0.05 & 1.32\%$\pm$0.09 \\
Guided-VAE \cite{ding2020guided}   
            & 1.85\%$\pm$0.08 & 1.60\%$\pm$0.08  & 1.49\%$\pm$0.06 \\            
Guided-$\beta$-TCVAE \cite{ding2020guided}   
            & 1.47\%$\pm$0.12  & \textbf{1.10\%$\pm$0.03}  & 1.31\%$\pm$0.06 \\
\hline
DC-VAE (Ours)   
            & \textbf{1.30\%$\pm$0.035 } & 1.27\%$\pm$0.037 & \textbf{1.29\%$\pm$0.034} \\
\bottomrule
\hline
\end{tabular}
}
\end{center}
\vspace{-3mm}
\end{table}

To show that our model learns a good representation, we measure the performance on the downstream MNIST classification task \cite{ding2020guided}. The VAE models were trained on MNIST dataset \cite{lecun2010mnist}. We feed input images into our VAE encoder and get the latent representation. Then we train a linear classifier on the latent representation to classify the classes of the input images. Results in Table \ref{tab:classification-mnist-methods} show that our model gives the lowest classification error in most cases. This experiment demonstrates that our model not only gains the ability to do faithful synthesis and reconstruction, but also gains better representation ability on the VAE side.

\begin{table}[h!]
    \centering
    \caption{PPL Comparison of on CelebA-HQ \cite{karras2018progressive}. }
    \label{tab:PPL}
    \scalebox{0.8}{
    \begin{tabular}{@{}lccccccc@{}}
        \hline
            \toprule
            \textbf{Method} & \textbf{Backbone} & \textbf{PPL Full}$\downarrow$  \\
            \midrule
            StyleALAE \cite{pidhorskyi2020adversarial} & StyleGAN \cite{karras2019style}  & 33.29 \\
            ProGAN \cite{karras2018progressive}  & ProGAN \cite{karras2018progressive} & 40.71 \\
            DC-VAE (ours) & ProGAN \cite{karras2018progressive} & \textbf{24.66}\\
            \bottomrule
            \hline
        \end{tabular}
        }
\end{table}

\begin{table}[h!]
    \centering
    \caption{\textbf{Reconstruction Comparison of on CelebA-HQ} \cite{karras2018progressive} validation set. We follow \cite{Johnson2016Perceptual} and measure perceptual distance in an relu$4\_3$ layer of a pretrained VGG network. $\downarrow$ means lower is better.}
    \label{tab:recon}
    \scalebox{0.8}{
    \begin{tabular}{@{}lccc@{}}
        \hline
            \toprule
            \textbf{Method} & \textbf{Backbone} & \begin{tabular}{c} \textbf{Pixel}$\downarrow$ \\ \textbf{Distance} \end{tabular} & \begin{tabular}{c} \textbf{Perceptual}$\downarrow$ \\ \textbf{Distance} \end{tabular} \\
            \midrule
            StyleALAE \cite{pidhorskyi2020adversarial} & StyleGAN \cite{karras2019style} & 0.117 & 40.40 \\
            DC-VAE (ours) & ProGAN \cite{karras2018progressive} & \textbf{0.072} & \textbf{38.63}\\
            \bottomrule
            \hline
        \end{tabular}
        }
        \vspace{-5mm}
\end{table}

\vspace{-2mm}
\section{Conclusion}
\vspace{-2mm}
\label{sec:conclusion}

In this paper, we have developed dual contradistinctive generative autoencoder (DC-VAE), a new framework that integrates an instance-level discriminative loss (InfoNCE) with a set-level adversarial loss (GAN) into a single variational autoencoder framework. 
Our experiments show state-of-the-art or competitive results in several tasks, including image synthesis, image reconstruction, representation learning for image interpolation, and representation learning for classification. DC-VAE is a general-purpose VAE model and it points to a encouraging direction that attains high-quality synthesis (decoding) and inference (encoding).

\section{Acknowledgment}
This work is funded by NSF IIS-1717431 and NSF IIS-1618477. Zhuowen Tu is also funded under Qualcomm Faculty Award. 

{\small
\bibliographystyle{ieee_fullname}
\bibliography{egbib}
}

\newpage
\appendix
\section{Appendix}
\label{sec:appendix}
\subsection{\bf Additional reconstruction results}
\begin{figure*}[htp]
    \vspace{-1mm}
    \begin{center}
        \includegraphics[width=0.8\linewidth]{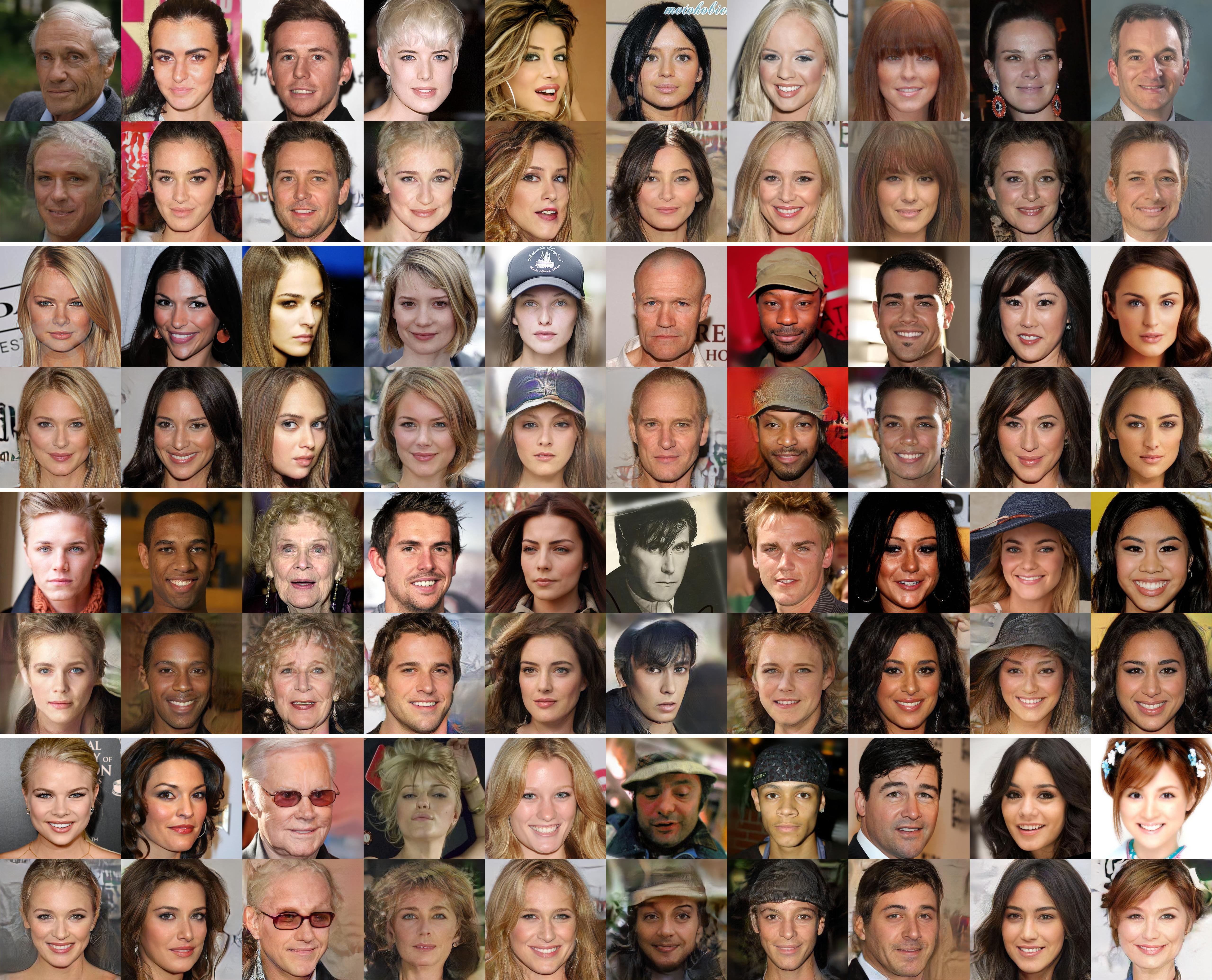}
        \includegraphics[width=0.8\linewidth]{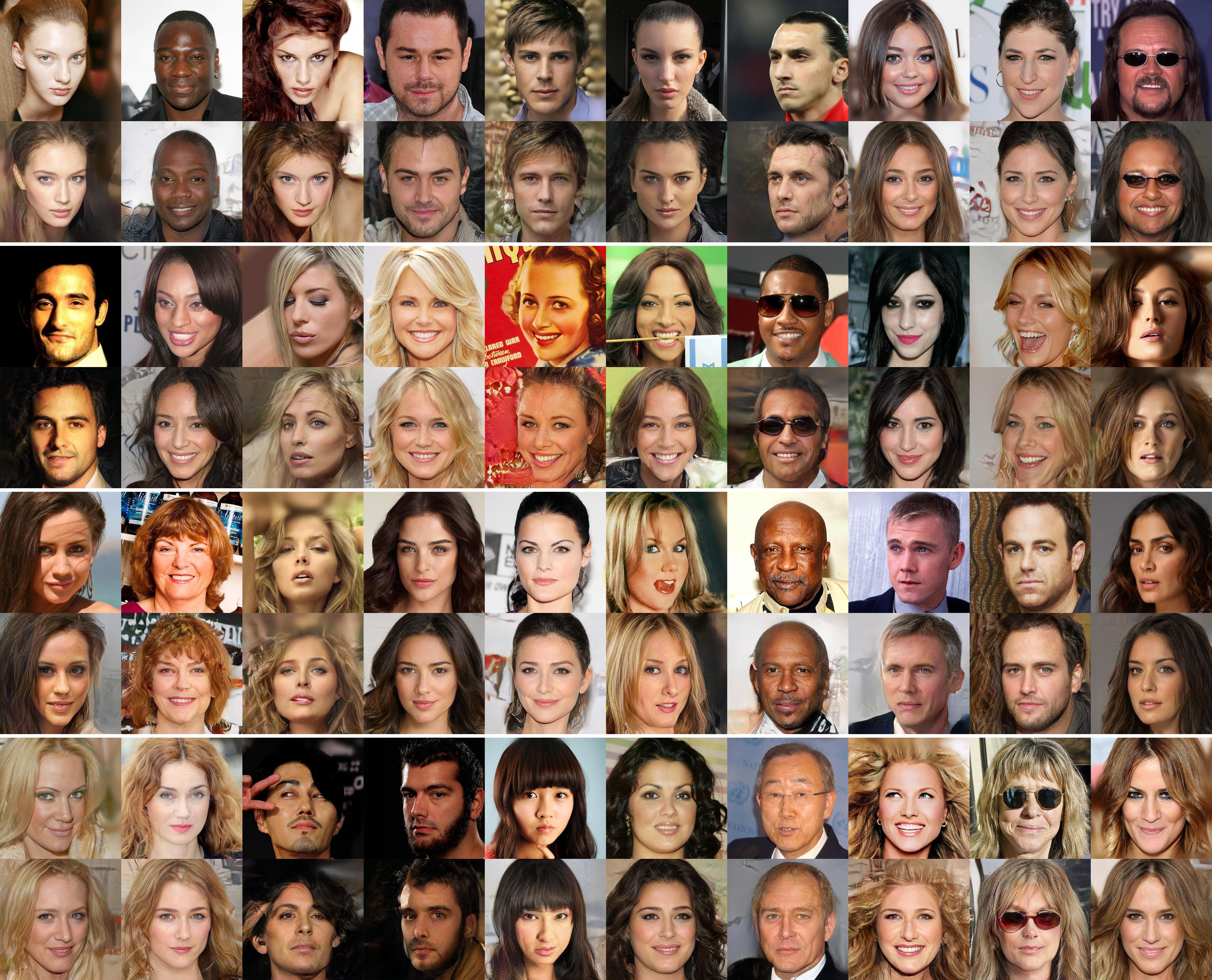}
      \caption{\small
        Additional CelebA-HQ \cite{karras2018progressive} reconstruction images (resolution \(512 \times 512\)) generated by DC-VAE (ours)  
      }
    \vspace{-3mm}
    \label{fig:celeba_large_grid}
    \end{center}
\end{figure*}

\begin{figure*}[h!]
    \vspace{-1mm}
    \begin{center}
        \includegraphics[width=0.95\linewidth]{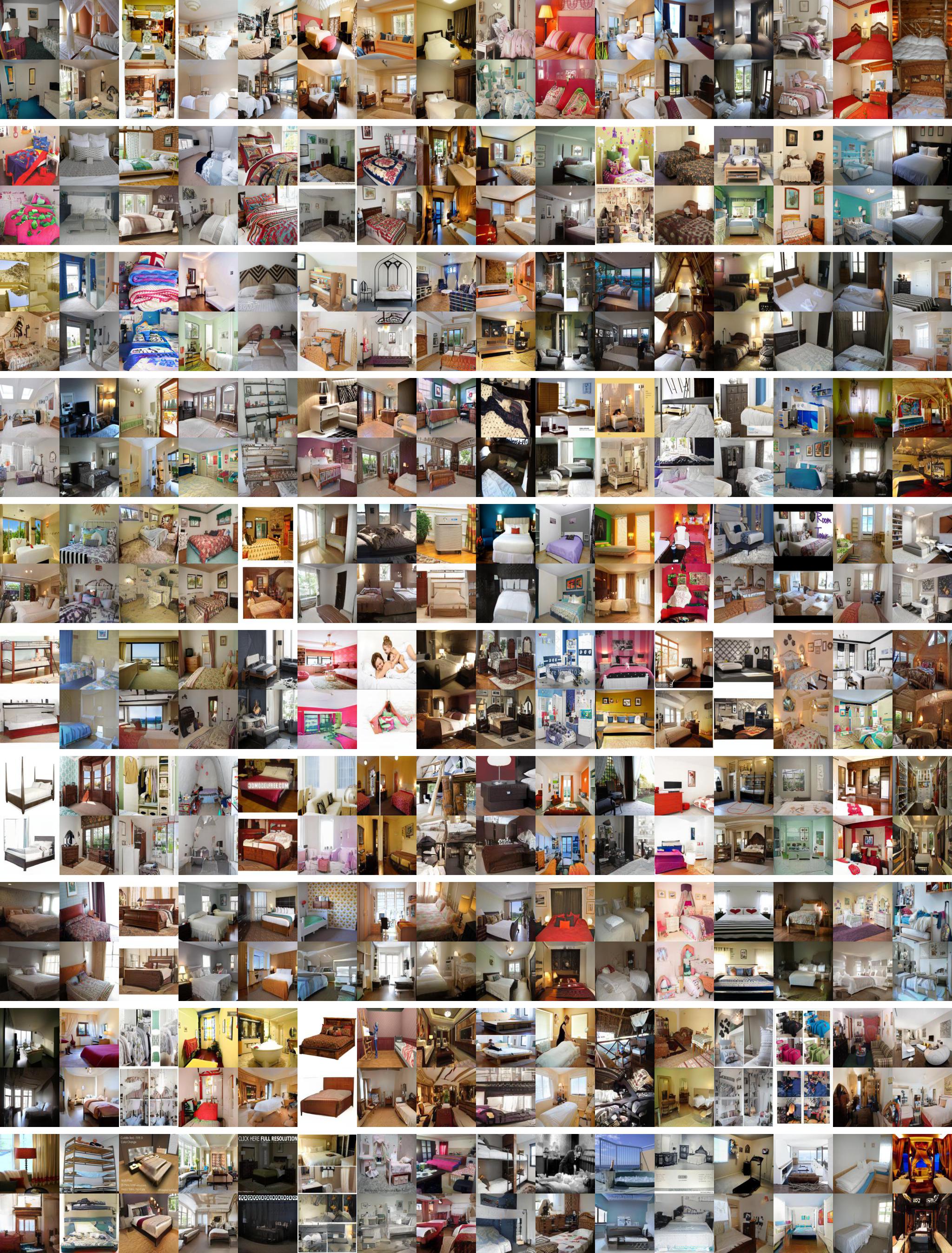}
      \caption{\small
        Additional LSUN Bedroom \cite{yu15lsun} reconstruction images (resolution \(128 \times 128\))
      }
    \vspace{-3mm}
    \label{fig:lsun_large_grid}
    \end{center}
\end{figure*}

In Figures \ref{fig:celeba_large_grid} and \ref{fig:lsun_large_grid} we show a large collection of additional recontruction images on the CelebA-HQ \cite{karras2018progressive} and LSUN Bedroom \cite{yu15lsun} datasets.

\subsection{\bf Smoothness of latent space}
\label{apsec:latent}
\begin{figure*}[h!]
    \begin{center}
        \includegraphics[width=0.7\linewidth]{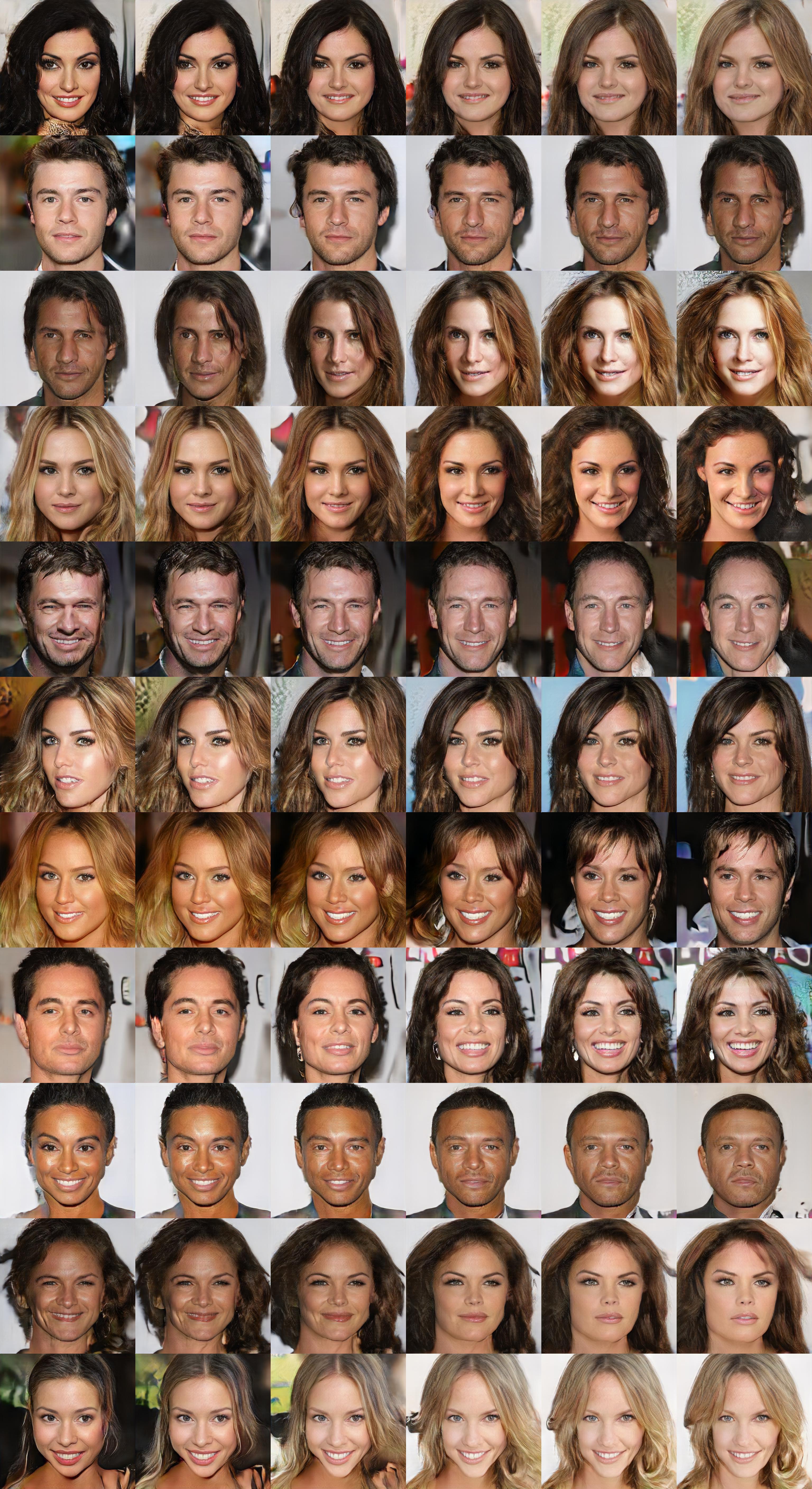}
        
        \caption{ \small Additional latent space interpolations on CelebA-HQ \cite{karras2018progressive} (resolution \(512 \times 512\))}
    \label{fig:celebahq_additional_interpolation}
    \end{center}
    
    \vspace{-7mm}
\end{figure*}

In this section we analyse the smoothness of the latent space learnt by DC-VAE. In Figure \ref{fig:celebahq_additional_interpolation} we show additional high resolution ($512 \times 512$) CelebA-HQ \cite{karras2018progressive} images generated by an evenly spaced linear blending between two latent vectors. In Fig. \ref{fig:CLB_editing} we show that DC-VAE is able to perform meaningful attribute editing on images while retaining the original identity. To perform image editing, we first need to compute the direction vector in the latent space that correspond to a desired attribute (e.g. has glasses, has blonde hair, is a woman, has facial hair). We compute these attribute direction vectors by selecting 20 images that have the attribute and 20 images that do not have the attribute, obtaining the corresponding pairs of 20 latent vectors, and calculating the difference of the mean. The results in Fig. \ref{fig:CLB_editing} show that these direction vectors can be added to a latent vector to add a diverse combination of desired image attributes while retaining the original identity of the individual.

\subsection{\bf Effect of negative samples}
\begin{figure}[H]
    \centering
    \includegraphics[width=\linewidth]{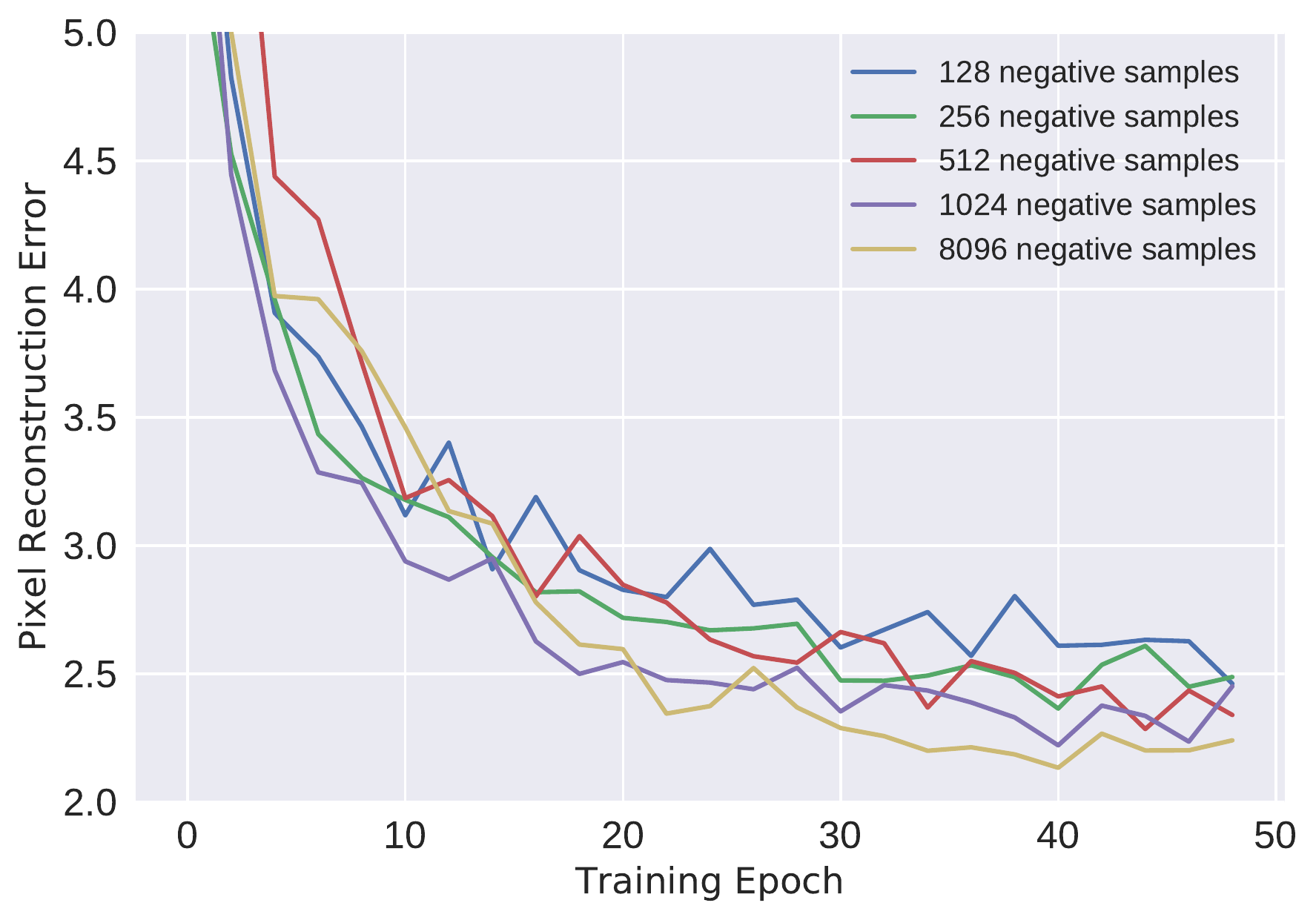}
    \vspace{-5mm}
    \caption{\small Pixel reconstruction error on CIFAR-10 \cite{cifar10} test set for varying number of negative samples}
    \label{fig:ablation_k}
\end{figure}

In this section we analyse the effect of varying the number of negative samples used for contrastive learning. Figure \ref{fig:ablation_k} shows the reconstruction error on the CIFAR-10 \cite{cifar10} test set as the negative samples is varied. We observe that a higher number of negative samples results in better reconstruction. We choose 8096 for all of our experiments because of memory constraints. 

\subsection{\bf Dataset details} 
CIFAR-10 comprises 50,000 training images and 10,000 test images
with a spatial resolution of $32 \times 32$. STL-10 is a similar dataset that contains 5,000 training images and 100,000 unlabeled images at $96 \times 96$ resolution. We follow the procedure in AutoGAN \cite{gong2019autogan} and resize the STL-10 images to $32 \times 32$. 
The CelebA dataset has 162,770 training images and 19,962 testing images, CelebA-HQ  contains 30,000 images of size $1024 \times 1024$, and LSUN Bedroom has approximately 3M images. For CelebA-HQ we split the dataset into 29,000 training images and 1,000 validation images following the method in \cite{huang2018introvae}. We resize all images progressively in these three datasets from  ($4 \times 4$) to ($512 \times 512$) for the progressive training.

\subsection{\bf Network architecture diagrams}
\begin{figure*}[!htp]
    \begin{center}
    \begin{tabular}{c c c}
              \includegraphics[width=0.2\linewidth]{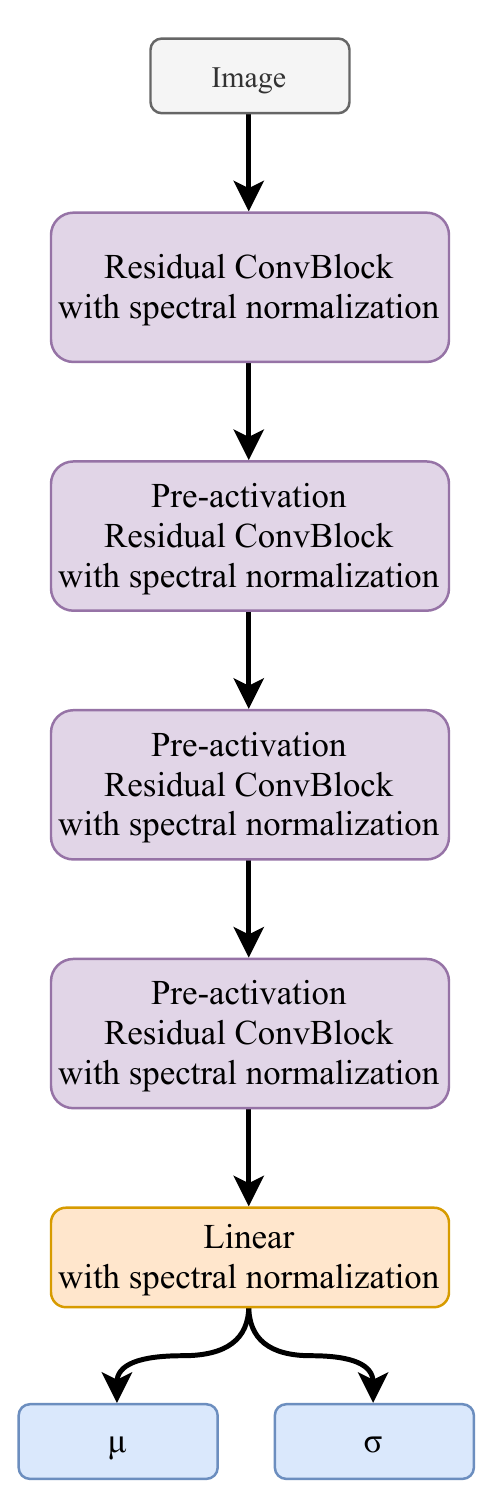} &
              \includegraphics[width=0.4\linewidth]{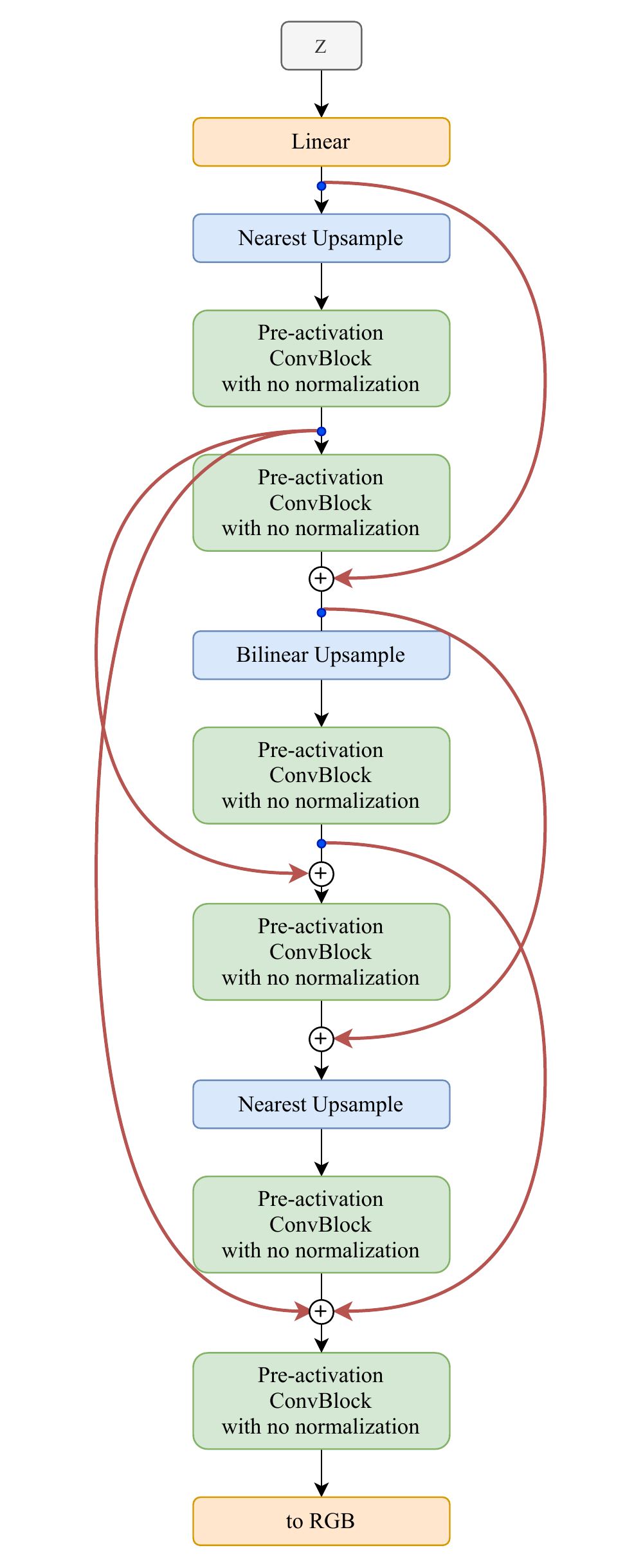} & 
              \includegraphics[width=0.375\linewidth]{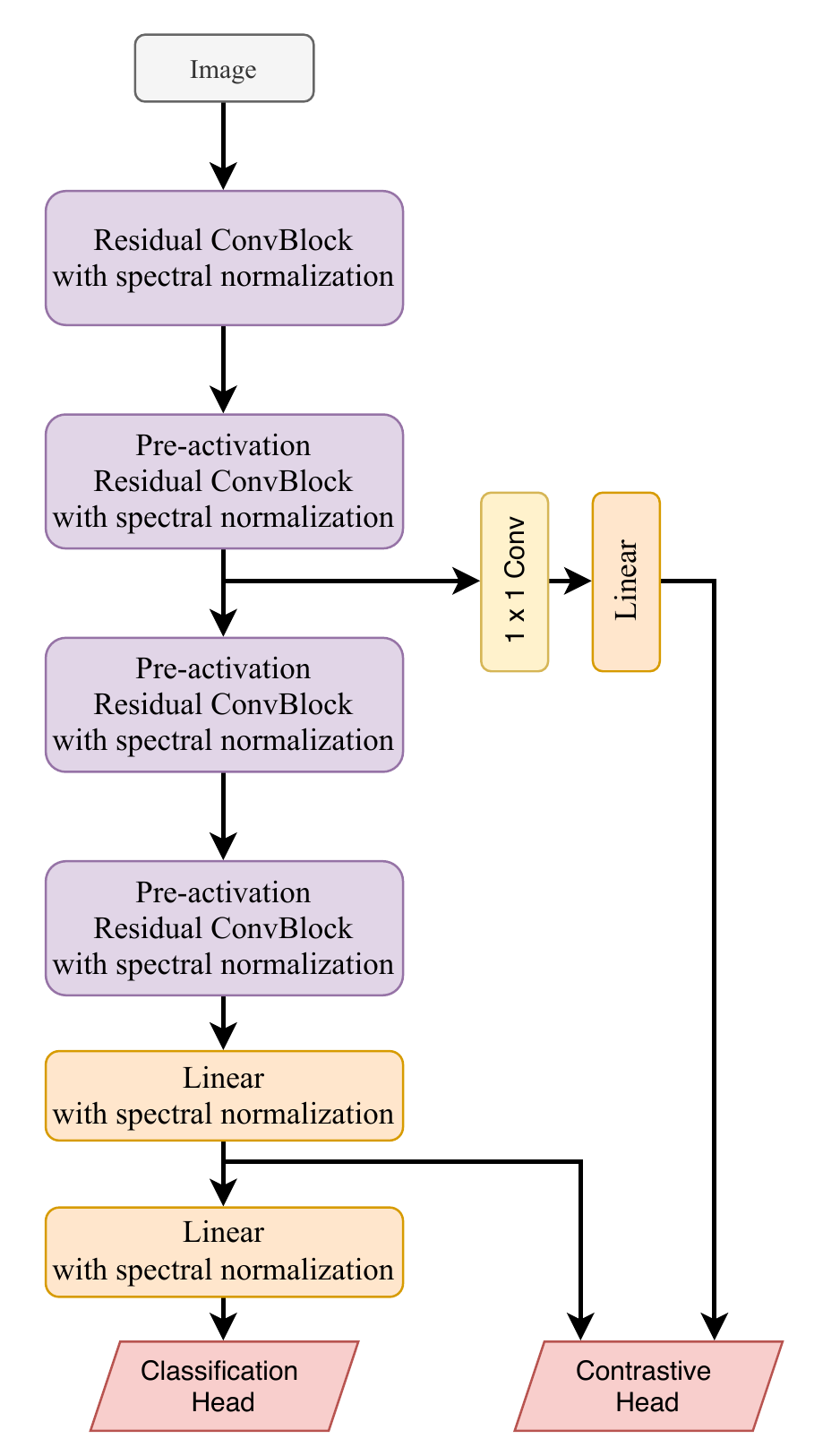} \\
            (a) Encoder & (b) Decoder & (c) Discriminator
        \end{tabular}
    \end{center}
    \caption{\small Network architecture of DC-VAE for resolution $32 \times 32$ for CIFAR-10 \cite{cifar10} and STL-10 \cite{STL10}. (a) is the Encoder. (b) is the Decoder. (c) is the Discriminator.}
    \label{fig:32_architecture}
\end{figure*}

In Figures \ref{fig:32_architecture} we show the detailed network architecture of DC-VAE for input resolutions of $32 \times 32$. 
Note that the comparison results shown in Figure \ref{fig:CF10_abl} and Table \ref{table:cifar_ablation} in the main paper, for VAE, VAE/GAN, VAE w/o GAN, and our proposed DC-VAE are all based on the same network architecture (shown in Figure \ref{fig:32_architecture} here), for a fair comparison.

The network architectures shown in Figure \ref{fig:32_architecture} are adapted closely from the networks discovered by \cite{gong2019autogan} through Neural Architecture Search. The DC-VAE developed in our paper is not tied to any particular CNN architecture. We choose the AutoGAN architecture \cite{gong2019autogan} to start with a strong baseline.
The decoder in Figure \ref{fig:32_architecture} matches the generator in \cite{gong2019autogan}. 
The encoder is built by modifying the output shape of the final linear layer in the discriminator of AutoGAN \cite{gong2019autogan} to match the latent dimension and adding spectral normalization. The discriminator is used both for classifying real/fake images, and contrastive learning. For each layer we choose, we first apply 1x1 convolution and a linear layer, and then use this feature as an input to the contrastive module. For experiments at $32 \times 32$, we pick two different positions:  the output of second residual conv block (lower level) and the output of the first linear layer (higher level).
For experiments on higher resolution datasets we use a Progressive GAN \cite{karras2018progressive} Generator and Discriminator as our backbone and apply similar modifications as described above.

\subsection{\bf Further details about the representation learning experiments}
As seen in Table 4 in the main paper, we show the representation capability of DC-VAE following the procedure outlined in \cite{ding2020guided}. We train our model on the MNIST dataset \cite{lecun2010mnist} and measure the transferability though a classification task on the latent embedding vector. Specifically, we first pretrain the DC-VAE model on the training split of the MNIST dataset. Following that we freeze the DC-VAE model and train a linear classifier that takes latent embedding vector as the input and predicts the class label of the original image.

\begin{figure*}[h!]
    \begin{tabular}{c|c}
         \includegraphics[width=0.5\linewidth]{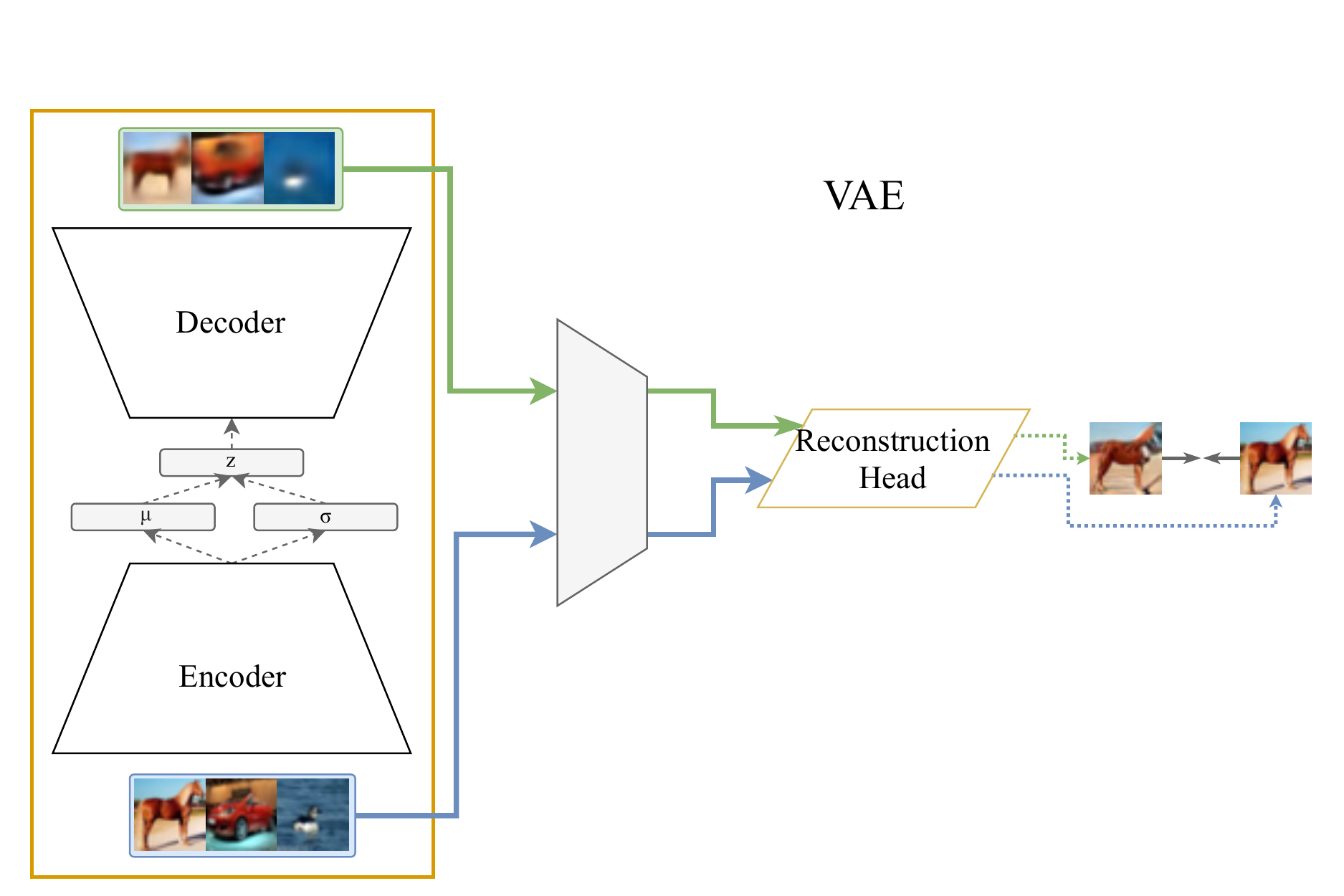}&
         \includegraphics[width=0.5\linewidth]{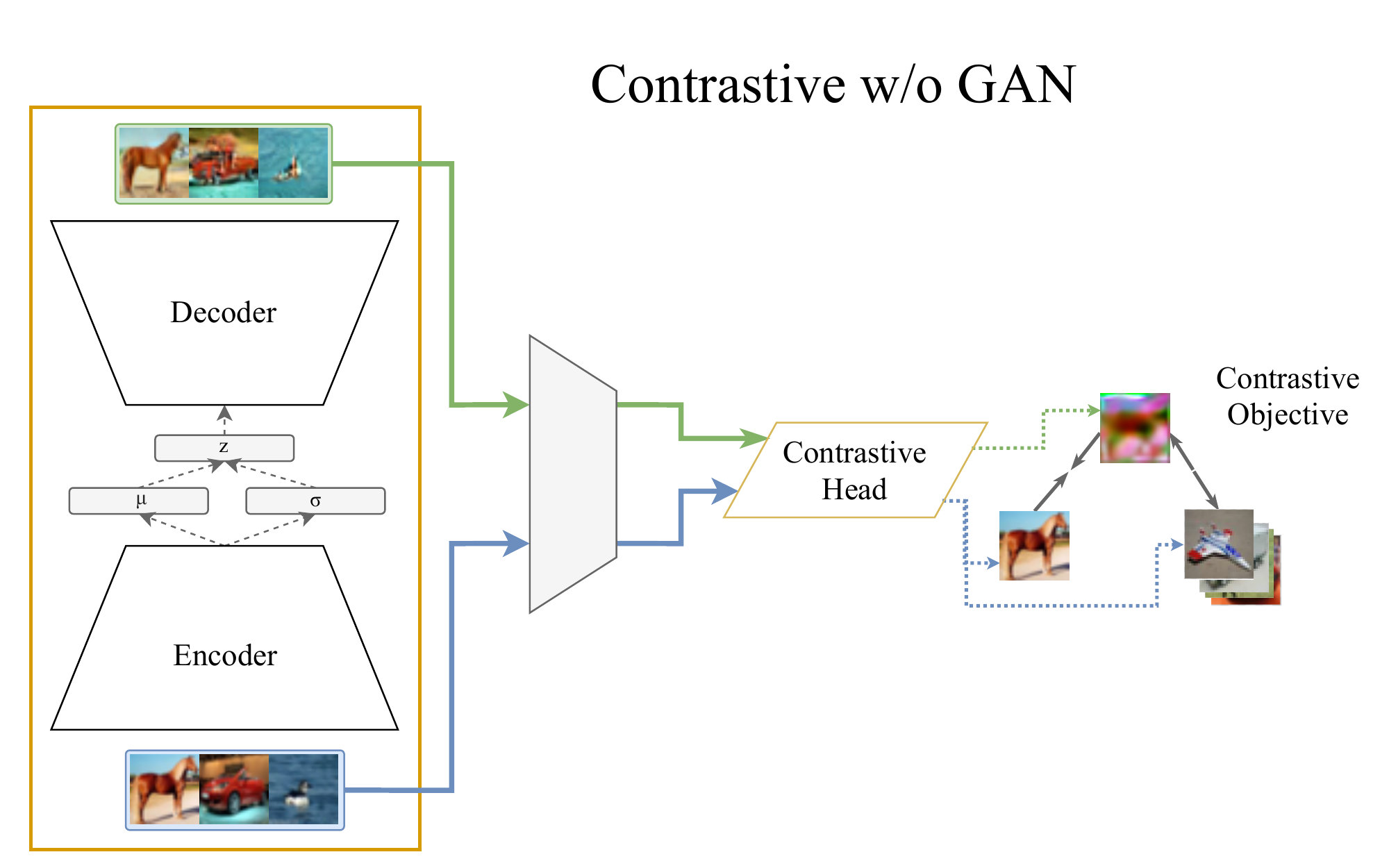}\\
         \hline
         \includegraphics[width=0.5\linewidth]{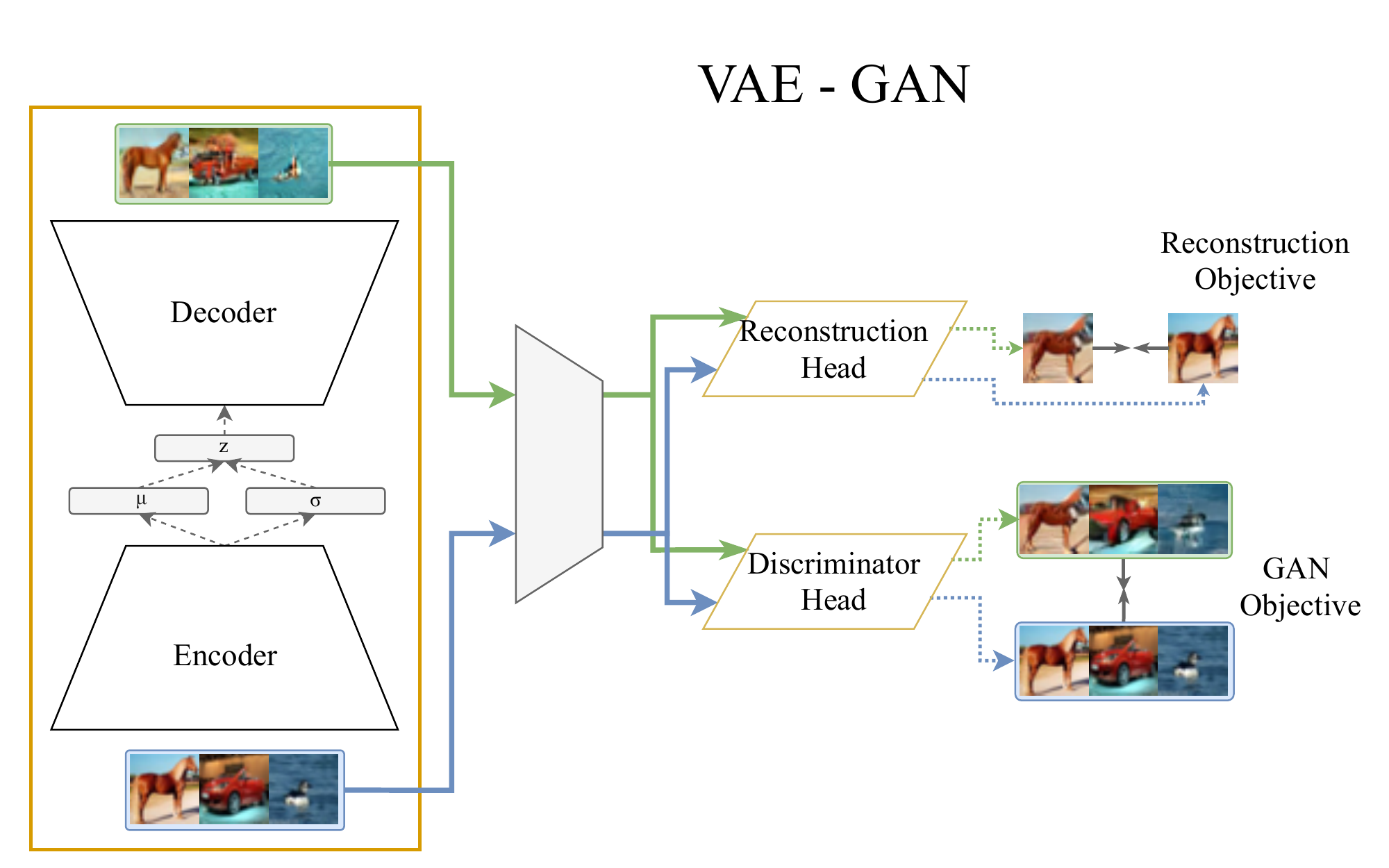}&
         \includegraphics[width=0.5\linewidth]{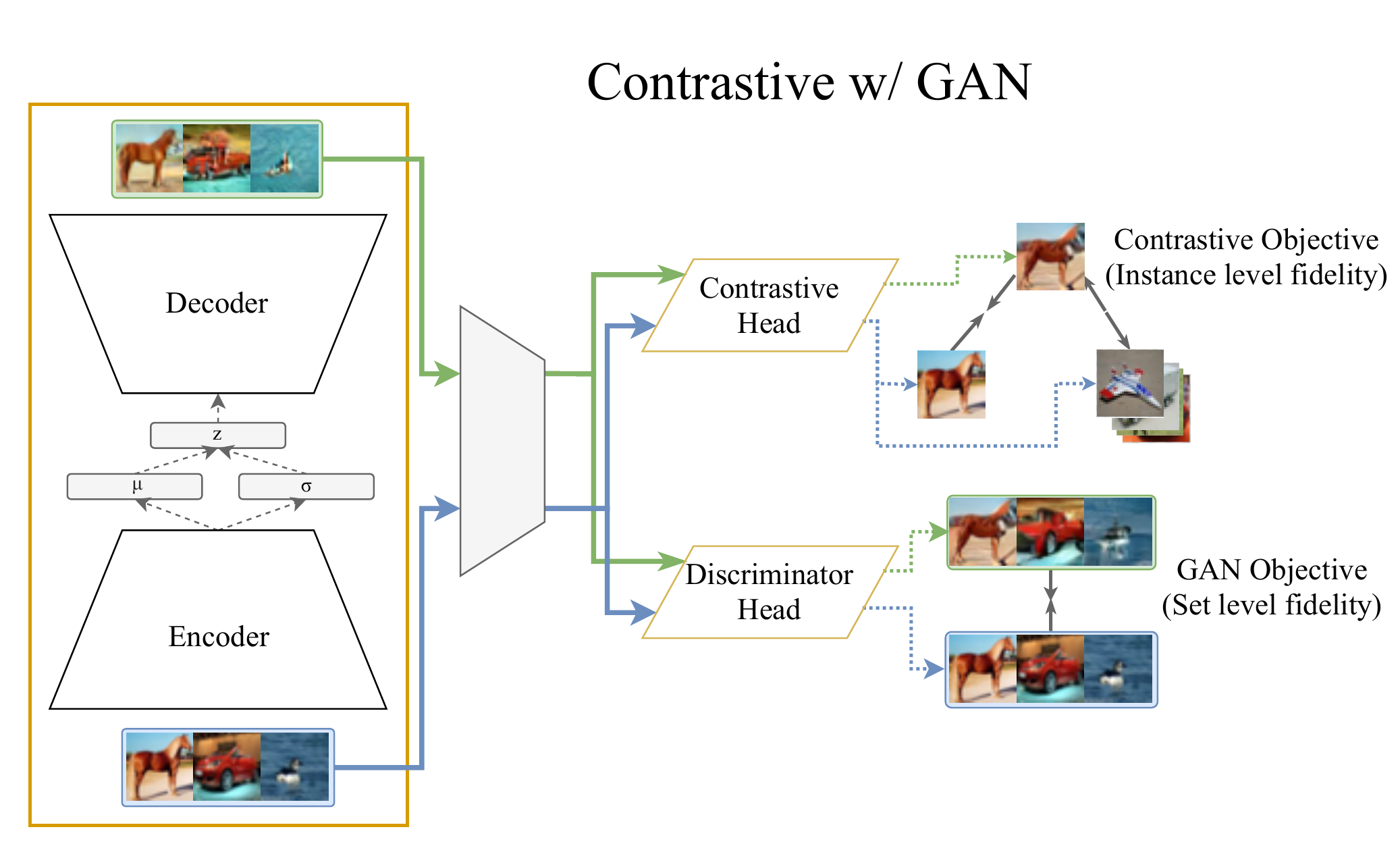}\\
    \end{tabular}
    \caption{ \small Visualization of the effect of adding each instance level and set level objectives. Table \ref{table:cifar_ablation} and Figure \ref{fig:CF10_abl} contain FID \cite{FID} results and qualitative comparisons on the CIFAR-10 \cite{cifar10} that correspond to these settings.}
    \label{fig:cifar_ablation}
\end{figure*}

\begin{figure*}[h!]
    \begin{center}
        \subfigure[{\small STL-10 Reconstructions generated by DC-VAE }]{\includegraphics[width=0.4\linewidth]{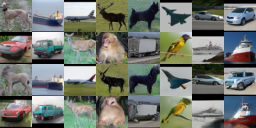}}
        \subfigure[{\small STL-10 Samples generated by DC-VAE }]{\includegraphics[width=0.4\linewidth]{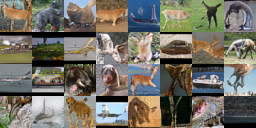}}
    \end{center}
    \vspace{-5mm}
    \caption{ \small DC-VAE reconstruction (a) and synthesis results (b) on STL-10 \cite{STL10} images (resolution \(32 \times 32\)). In (a) the top two rows are input images and the bottom two rows are the corresponding reconstruction images. }
    \label{fig:stl10_images}
\end{figure*}

\subsection{\bf Evaluation details}
In Tables \ref{table:cifar_ablation} and \ref{tab:recon} the perceptual distance is computed as the average MSE distance of the features extracted by a pretrained VGG-16 network. We borrow from  \cite{Johnson2016Perceptual} and use the activation of the relu4\_3 layer. For computing the FID scores we follow the standard practice (\cite{huang2018introvae}, \cite{pidhorskyi2020adversarial}) and use 50,000 generated images. In Table \ref{table:celebahq} we use the \(256 \times 256\) version of DC-VAE model trained on CelebA-HQ \cite{karras2018progressive} for a fair comparison with other methods which are trained at the same resolution.

\begin{figure*}[ht!]
    \centering
    \includegraphics[width=1.0\linewidth]{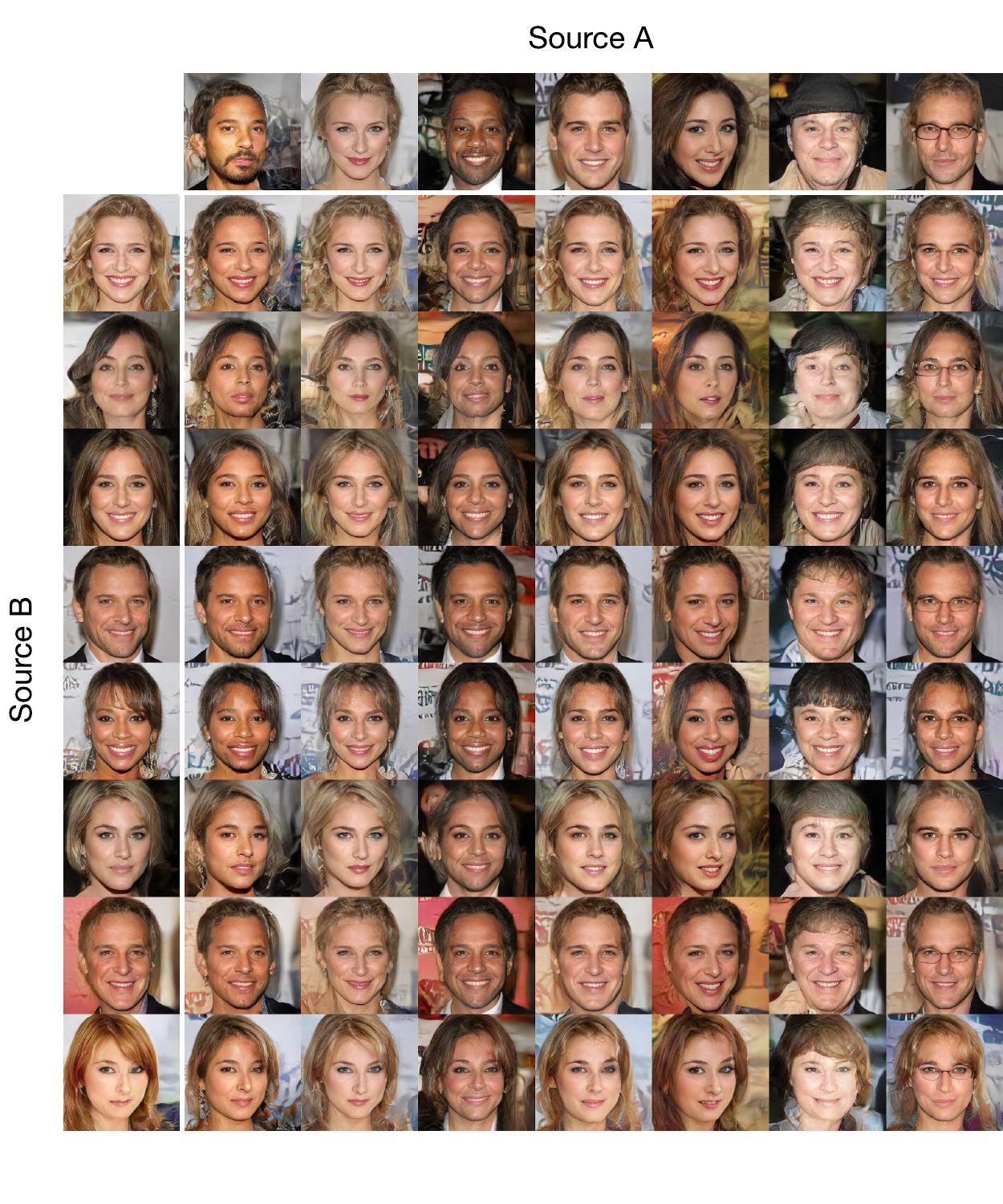}
    \vspace{-10mm}
    \caption{Latent Mixing results on CelebA-HQ \cite{karras2018progressive}. Each combined image in the grid is generated by replacing an arbitrary subset of Source A latent with the corresponding Source B latent.}
    \label{fig:latent_mixing}
\end{figure*}

\begin{figure*}[h!]
    \vspace{-1mm}
    \begin{center}
        \includegraphics[width=\linewidth]{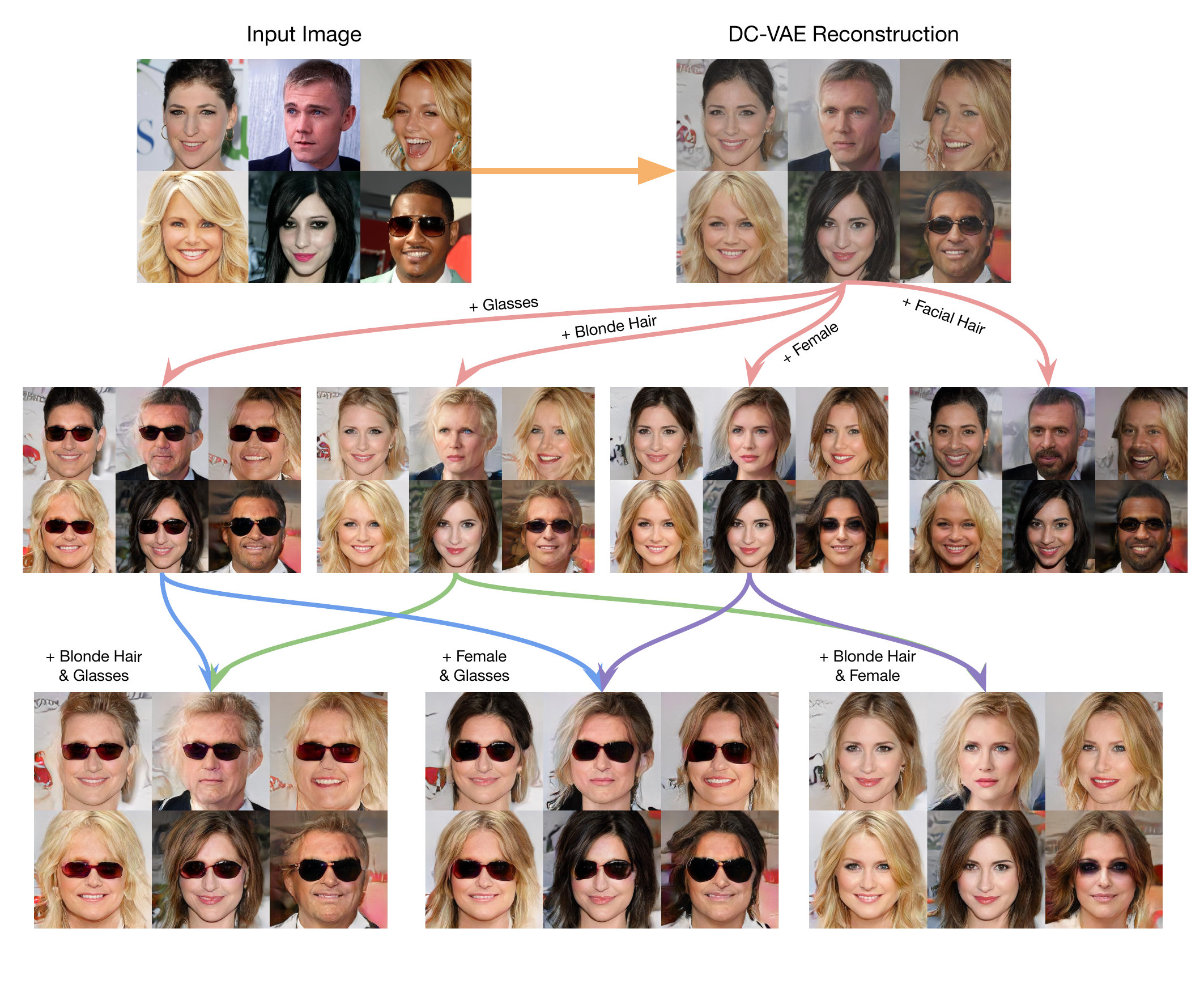}
      \vspace{-10mm}
      \caption{\small
        Additional image editing on CelebA-HQ \cite{karras2018progressive} reconstruction images (resolution \(512 \times 512\))
      }
    
    \label{fig:celebahq_edit_reconstruction}
    \end{center}
\end{figure*}

\end{document}